\documentclass[journal]{IEEEtran} 
\IEEEoverridecommandlockouts
\usepackage{xcolor}
\usepackage{tablefootnote}
\usepackage[
backend=biber,
style=ieee,
sorting=none
]{biblatex}
\usepackage{amsmath,amssymb,amsfonts,mathtools}
\usepackage{algorithmic}
\usepackage{bm}
\usepackage{makecell}
\usepackage{graphicx}
\usepackage[utf8]{inputenc}
\usepackage[english]{babel}
\newtheorem{myprop}{\textbf{Property}}
\newtheorem{mydef}{\textbf{Definition}}
\newtheorem{myassumption}{\textbf{Assumption}}
\newtheorem{mylemma}{\textbf{Lemma}}
\usepackage{tabularx,booktabs}
\newcolumntype{C}{>{\centering\arraybackslash}X} 
\usepackage{lipsum}
\usepackage{authblk}
\usepackage{textcomp}
\usepackage{units}

\DeclareMathOperator*{\argmin}{arg\,min}

\usepackage{subcaption}
\usepackage[export]{adjustbox}
\usepackage[export]{adjustbox}

\usepackage[ruled, lined, linesnumbered, commentsnumbered, longend]{algorithm2e}
\SetKwInput{KwInput}{Input}                
\SetKwInput{KwOutput}{Output}
\newcommand\mymathbb[1]
  { {\rm I\mathchoice{\hspace*{-2pt}}{\hspace*{-2pt}}
    {\hspace*{-1.75pt}}{\hspace*{-1.7pt}}#1}   }
\newcommand{\R}{\mymathbb R}
\newcommand{\N}{\mymathbb N}

\begin{document}
\title{Adaptive Safety-critical Control with Uncertainty Estimation for Human-robot Collaboration\\
\thanks{The authors are with the Centre for Intelligent Autonomous Manufacturing Systems (i-AMS), School of Electronics, Electrical Engineering and Computer Science, Queen’s University Belfast, Belfast BT9 5AG, U.K. E-mail: dzhang07, m.van, s.mcloone, smcilvanna01, ysun32@qub.ac.uk.}}

\author{Dianhao Zhang, Mien Van, Stephen Mcllvanna, Yuzhu Sun, Seán McLoone\IEEEmembership{, Senior Member,~IEEE}}
\maketitle
\begin{abstract}
   In advanced manufacturing, strict safety guarantees are required to allow humans and robots to work together in a shared workspace. One of the challenges in this application field is the variety and unpredictability of human behavior, leading to potential dangers for human coworkers. This paper presents a novel control framework by adopting safety-critical control and uncertainty estimation for human-robot collaboration. Additionally, to select the shortest path during collaboration, a novel quadratic penalty method is presented. The innovation of the proposed approach is that the proposed controller will prevent the robot from violating any safety constraints even in cases where humans move accidentally in a collaboration task. This is implemented by the combination of a time-varying integral barrier Lyapunov function (TVIBLF) and an adaptive exponential control barrier function (AECBF) to achieve a flexible mode switch between path tracking and collision avoidance with guaranteed closed-loop system stability. The performance of our approach is demonstrated in simulation studies on a 7-DOF robot manipulator. Additionally, a comparison between the tasks involving static and dynamic targets is provided. 
\end{abstract}
{
\def\abstractname{Note to Practitioners}
\begin{abstract}
This research addresses the need to improve the safety of robots interacting with humans when performing collaborative tasks. Existing safety-critical control (SCC) approaches do not adequately monitor and continuously limit the state of the robot in Cartesian space, which results in a risk of injury to human operators if there is unexpected behavior during collaboration. Additionally, existing SCC approaches only consider system uncertainty for a single task (i.e. path tracking only or collision avoidance only). These problems limit the applicability of SCC techniques to manufacturing cobots. 

We address these problems by developing a controller that accounts for uncertainty in robot dynamics, guarantees that the robot end-effector remains within a constrained task space, and continuously modifies its motion in real-time to avoid dynamic obstacles that violate this space. We employ a machine learning approach to estimate the unknown uncertainties in real-time, allowing them to be incorporated within the controller design. The designed controller selects the shortest path for collision avoidance at each sample instant in order to minimize the total motion of the robot.

\end{abstract}
}

\section{Introduction}
Modern control applications prioritize safety, and as system complexity rises, it becomes increasingly important to encode safety features when designing controllers properly.  Real-time human-robot collaboration (HRC) is an example of a rapidly growing application where safe control synthesis is critical.  To ensure human-worker safety in HRC enivornments, the International Organization for Standardization (ISO) has published safety regulations ISO 10218-2 and ISO/TS 15066  that specify limits on the operating speed, power, and force of robots working in such environments \cite{fryman2012safety}. However, guaranteeing the safety of humans in HRC is challenging, especially in some unexpected emergency situations such as unexpected human behavior occurring or humans moving beyond the workspace boundary by accident.  

To that end, we seek to develop a safety-critical-based controller for HRC. Safety-critical control (SCC) has been widely employed in industrial HRC tasks which require both accuracy of control and guaranteed safety of human operators, driven by its ability to bind the states and the system error so as to achieve adaptive constrained control. It has the benefits of guaranteeing stability and safety. The barrier Lyapunov function and control barrier function are famous tools in safety-critical control. We apply these tools to solve the HRC tasks involving path tracking and collision avoidance in this work. 

For path tracking problems, the barrier Lyapunov function (BLF) is an important tool in SCC, which is developed by reshaping the structure of the control Lyapunov function (CLF) with the barrier function's characteristics \cite{ngo2005integrator}. Such a function grows to infinity whenever the target parameter approaches the defined limits. BLF and its extensions (e.g. sBLF \cite{nohooji2018neural}, log-type BLF \cite{liu2018adaptive}, tan-type BLF \cite{tang2013tangent} and IBLF \cite{he2020admittance}) are widely employed in the area of autonomous vessels, wind turbines, admittance control, and industrial manufacturing \cite{jin2016fault, habibi2017constrained, he2020admittance}. In contrast to existing approaches, we propose using a time-varying IBLF (TVIBLF) to perform the path-tracking task in HRC. This allows us to constrain the system output directly rather than the tracking error, which may be more convenient for controller design.

As for collision avoidance, the control barrier function is a subchannel of SCC, which was first proposed by  Wieland et al. \cite{wieland2007constructive} as an extension of the barrier function. The objective is to simultaneously guarantee the asymptotic stability of the input signal while constraining it in a safe setting. High-order CBF (HOCBF) and exponential CBF (ECBF) extensions are proposed in \cite{son2019safety} to solve higher-order relative degree problems. In this work, a position state modification is implemented based on ECBFs with relative degree two and neural networks to achieve smooth and path-saving collision avoidance solved by the quadratic penalty programming method. The proposed adaptive ECBF (AECBF) safety filter seeks to minimize the total movement during collision avoidance. The idea behind it is to take advantage of a feedforward neural network (FNN) to estimate the robot's Cartesian state. Then, a penalty term is generated reflecting the distance between agents.  



In order to achieve accurate positioning in manipulator control, the system uncertainty needs to be considered. Fuzzy logic can be used to compensate for dynamics uncertainties, and has been widely employed in manipulator control tasks \cite{dimeas2015human, kahraman2020fuzzy}. These works have verified that fuzzy systems can model the uncertainty as a nonlinear function of the joint position error and its time derivative for more accurate positioning during control. Neural networks are also widely used in uncertainty estimation problems \cite{wu2015global, he2015neural}. They can also be used in combination with fuzzy logic (via fuzzy neural network models), as demonstrated in \cite{he2017adaptive}.
To achieve a safe HRC system with unmodeled uncertainty,
%
%
we propose a Radial Basis Function Neural Network (RBFNN) based human-robot interaction controller to approximate the system uncertainty at each time instance, which helps to improve the performance of path tracking and collision avoidance. The improvement achieved with this approach is demonstrated with ablation studies in different designed experiments.  


%
%
In summary, this paper makes contributions to both the control theory and robot application domains as follows:
\begin{itemize}
\item Compared with existing ECBF based control designs, which require full knowledge of the dynamics of a system \cite{2021Control} or have not considered the effects of uncertainty \cite{9777251}, this paper integrates ECBF and an NN estimator to eliminate the effects of uncertainty on the system dynamics for the first time.

\item Compared with the existing approaches that consider the use of IBLF or EBCF for safety critical control \cite{9353988}, this paper uses a combination of IBLF and ECBF to ensure that constraints on system states and obstacle avoidance can both be achieved. It should be noted that the integration of IBLF and ECBF is not straightforward, and it is very challenging to establish the stability of the closed-loop IBLF-ECBF. This paper establishes the stability for the closed-loop IBLF-ECBF for the first time.

\item With existing ECBF-based approaches for collision avoidance \cite{desai2022clf, jankovic2021collision}, the robot does not consider the optimal path to take in executing the collision avoidance maneuver. In this paper, we propose a new adaptive ECBF (ACBF), which adds a penalty term based on an MLP neural network prediction of the task space position of the end effector to achieve the minimum movement to execute collision avoidance.
\end{itemize}

\begin{itemize}
    \item In the context of human-robot collaboration, this is the first time that achieving enhanced safety through the integration of IBLF and ECBFs has been considered.
\end{itemize}

\section{Related Works}
\subsection{Control Lyapunov Function}
Lyapunov methods are a powerful tool for certifying the stability properties of nonlinear systems \cite{khalil2015nonlinear}. The use of Control Lyapunov Functions to synthesize stabilizing controllers for robotic platforms has become increasingly popular \cite{galloway2015torque}, often via quadratic programs (QPs) \cite{ames2013towards}. Despite the optimization-based formulation of these controllers, they often fail to achieve long-term optimal behavior. This deficiency arises due to the fact that the cost of these optimization problems fails to incorporate the future behavior of the system but is instead point-wise optimal \cite{freeman1996inverse}.

\subsection{Barrier Lyapunov Function}
As a means to extend the safety guarantees beyond the boundary of the set, there have been a variety of approaches that can be best described as “Lyapunov-like”. That is, Lyapunov functions yield invariant level sets so if these level sets are contained in the safe set one can guarantee safety. Importantly, these conditions can be applied over the entire set and not just on the boundary. Due to actual physical device limitations \cite{li2016adaptive}, system performance, and safety requirements \cite{he2017vibration}, output or states in most systems should be constrained in practice \cite{sun2009analysis}. Therefore, it is considered significant to maintain the system’s outputs in the desired constraints \cite{liu2017barrier}. For a nonlinear system of manipulators, output constraints can be regarded as position constraints. In recent years, the barrier Lyapunov function (BLF) has been proposed for solving output-constrained issues in complex systems \cite{zhang2018adaptive}. In \cite{tee2011control}, an asymmetric time-varying BLF is employed in strict feedback nonlinear systems to ensure the time-varying output constraints. In \cite{liu2018adaptive}, an adaptive control scheme is developed for nonlinear stochastic systems with unknown parameters. All the states of the systems are required to be constrained in bounded compact sets with log-type BLF. In \cite{zhou2016adaptive}, the output constraint problem of uncertain nonstrict-feedback systems is handled by utilizing a BLF. In \cite{zhou2016adaptive}, a tan-type BLF is used to maintain the output within constraints under systematic control design for strict-feedback nonlinear systems. In \cite{jin2016fault}, the tan-type BLF is incorporated with a novel fault-tolerant leader–follower formation control scheme to ensure the angle constraints. Compared with the conventional log- and tan-type BLFs, controllers with a novel integral BLF (IBLF) can constrain state signals directly, rather than error signals \cite{li2017adaptive}. From the engineering point of view, the initial states of robots can be relaxed to the whole constrained space.

\subsection{Control Barrier Function}
The notion of a barrier certificate was extended to a ``control” version to yield the first definition of a ``control barrier function” \cite{wieland2007constructive}. Although control barrier functions (CBFs) \cite{wieland2007constructive} are reminiscent of potential functions, CBF-based control design guarantees adherence to constraints. Knowledge about the system dynamics is combined with the CBFs in order to render a constraint-admissible subset of the state space-controlled invariant. In \cite{wills2004barrier}, this is achieved via model predictive control with CBFs. For improved real-time capabilities and to include performance specifications, Ames \emph{et al.} \cite{ames2019control} suggest a combination of CBFs with control Lyapunov functions (CLFs) via quadratic programming for cruise control. The approach is also successfully applied to bipedal robotic walking \cite{ames2014control} and for pendulum, control \cite{wu2015safety}. The existing control strategies require, however, a joint control design for task execution and constraint enforcement. As there are several well-established control schemes to achieve goal-directed behavior for robotic systems, a combination of these methods with CBFs is desirable. Additionally, scenarios involving interaction with humans require the enforcement of multiple and time-varying constraints, which have not been formally discussed for CBFs so far. In this work, BLFs are used to take account of internal constraints on links, motors, path tracking, etc., while CBFs are used to deal with external constraints (obstacles, etc.).

\section{Methodology}
\begin{figure*}
    \centering
    \includegraphics[width=\textwidth]{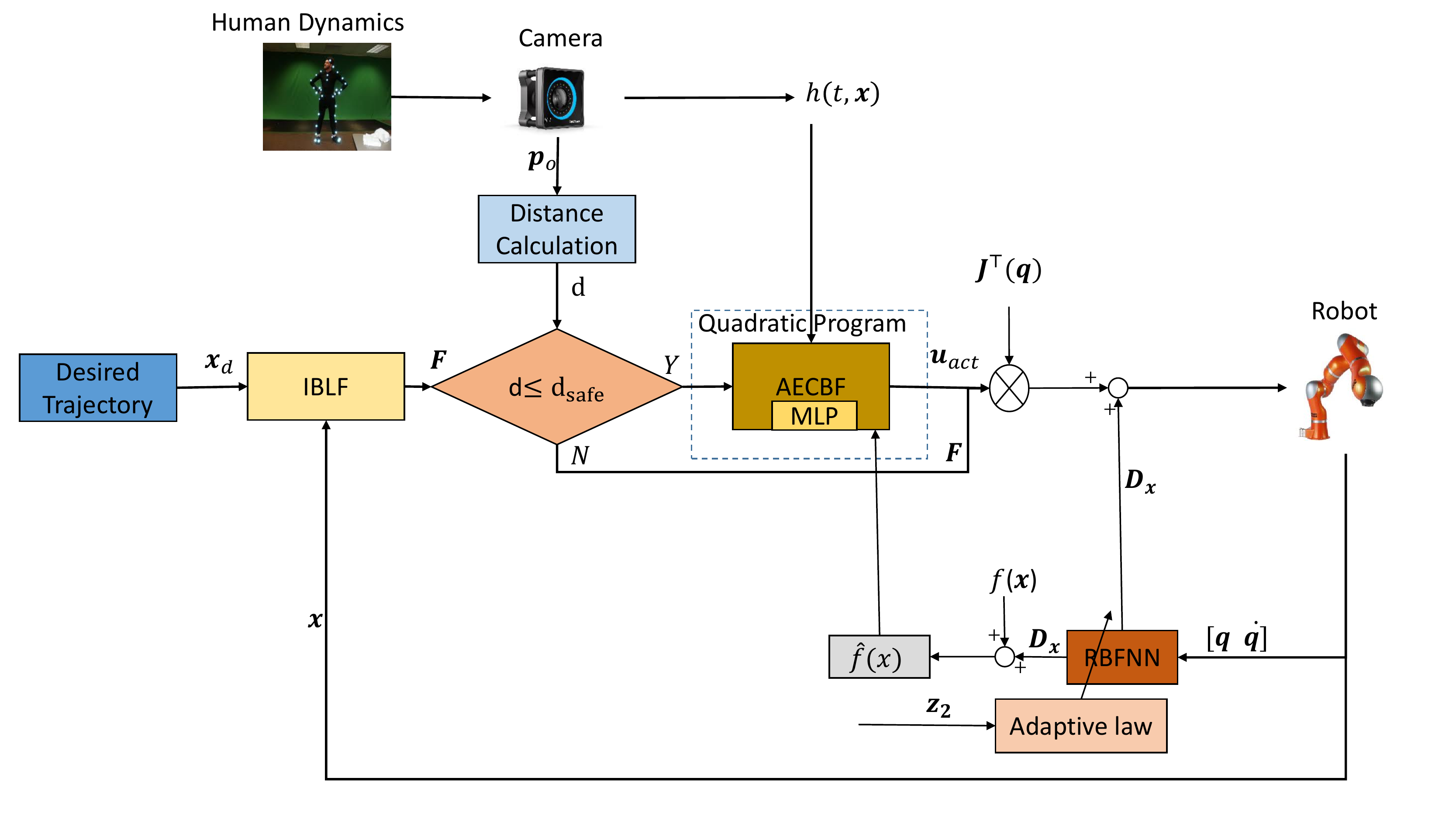}
    \caption{Architecture of the proposed control scheme.}
    \label{fig:architechture}
\end{figure*}

The overall design of the collision-free HRC system is shown in Fig. \ref{fig:architechture}. The proposed HRC system consists of motion capture, decision-making, path planning, uncertainty approximation, and robot controller modules. The motion capture module relies on cameras to reflect the position of the human. In this work, we use two obstacles in sphere format representing the body components of a human. The decision-making module is based on a metric reflecting the minimum distance between agents under the supervision of the perception system. The nominal controller for path tracking takes advantage of TVIBLF, while the safety of the human operator is guaranteed by the combination of the decision-making module and an AECBF-based safety filter. An RBFNN-based friction estimator is employed and propagates the estimated uncertainty to both the TVIBLF model and the AECBF model. In our work, a robot follows a desired trajectory with continuously constrained output and is required to generate a collision-free trajectory with the minimum distance. Two experiments are designed to test the performance of the proposed model. These involve the robot interacting with static and dynamic obstacles, respectively. 

\section{Formulation of the Optimization Problem}
\subsection{Dynamic Model}
Considering the joint friction model,  an $m$-degrees-of-freedom manipulator with equations of motion is defined as
\begin{align}
 \label{eq:dynamic}
     &\bm{M}(\bm{q})\Ddot{\bm{q}} + \bm{C}(\bm{q}, \dot{\bm{q}})\dot{\bm{q}} + \bm{g}(\bm{q}) + \bm{D}(\bm{q}, \dot{\bm{q}})  = \bm{\tau},\\
     &\bm{x} = f_{\mathrm{fwd}}(\bm{q}),
 \end{align}
where $\bm{q} \in \R^m$ and $\dot{\bm{q}}\in \R^m$ are the joint position and velocity of the robot, $\bm{M}(\bm{q}) \in \R^{m\times m}$ is a symmetric positive definite mass matrix, $\bm{C}(\bm{q}, \dot{\bm{q}}) \in \R^{m\times m}$ denotes Coriolis and centrifugal forces terms, $\bm{g}(\bm{q})\in \R^m$ is a vector of gravitational forces, $\bm{D}(\bm{q}, \dot{\bm{q}})\in \R^m$ is a vector of uncertainty terms associated with the manipulator joints, including unmodeled friction dynamics, the impact of environmental disturbances, and inaccuracies in the model parameters. We consider the joint torque $\bm{\tau} \in \mathbb{R}^m$ as the system input. The output of the system $\bm{x}_i \in \mathbb{R}^n$ is the position of the $i^{\text{th}}$ joint of the manipulator in Cartesian space. 

Denoting $f_{\mathrm{fwd}}: \R^m\rightarrow \R^n$ as the kinematic transformation from the joint angle position $\bm{q}$ to the Cartesian position $\bm{x} = f_{\mathrm{fwd}}(\bm{q})$, we have
\begin{align}
\label{eq:xdot}
\begin{cases}
    \dot{\bm{x}} ={}& \bm{J}(\bm{q})\dot{\bm{q}}\\
    \Ddot{\bm{x}} ={}& \bm{J}(\bm{q})\Ddot{\bm{q}} + \dot{\bm{J}}(\bm{q})\dot{\bm{q}}
\end{cases},
\end{align}
where $\bm{J}=\bm{J}(\bm{q}) = \frac{\partial f_{fwd}}{\partial \bm{q}}$ is the Jacobian matrix.
By substituting (\ref{eq:xdot}) into (\ref{eq:dynamic}), the Cartesian robotic system dynamics can be expressed as 
\begin{align}
\label{eq:cartesiandy}
    \bm{M_x}(\bm{q})\Ddot{\bm{x}} + \bm{C_x}(\bm{q}, \dot{\bm{q}})\dot{\bm{x}} + \bm{g_x}(\bm{q}) + \bm{D_x}(\dot{\bm{q}})= \bm{F},
\end{align}
where $\bm{M_x}(\bm{q}) = \bm{J}^{\dagger \top}\bm{M}(\bm{q})\bm{J}^{\dagger}$, $\bm{C_x}(\bm{q}, \dot{\bm{q}}) = J^{\dagger \top}(\bm{C}(\bm{q}, \dot{\bm{q}})-\bm{M}(\bm{q})\bm{J}^{\dagger}\dot{\bm{J}})\bm{J}^{\dagger}$, $\bm{g_x}(\bm{q}) = \bm{J}^{\dagger\top}\bm{g}(\bm{q})$, $\bm{D_x}(\bm{q}, \dot{\bm{q}}) = \bm{J}^{\dagger}\bm{D}(\bm{q}, \dot{\bm{q}})$ and $\bm{F} = \bm{J}^{\dagger}\bm{\tau}$ is the vector of generated control forces. $\bm{J}^{\dagger}(\bm{q})$ denotes the pseudo-inverse of $\bm{J}(\bm{q})$, that is:
\begin{align}
    \bm{J}^{\dagger}(\bm{q}) = (\bm{J}^{\top}(\bm{q})\bm{J}(\bm{q}))^{-1}\bm{J}^{\top}(\bm{q}).
\end{align}
For notational simplicity, $\bm{M_x}$, $\bm{C_x}$, $\bm{g_x}$ and $\bm{D_x}$ will be used to denote $\bm{M_x}(\bm{q})$, $\bm{C_x}(\bm{q}, \dot{\bm{q}})$, $\bm{g_x}(\bm{q})$ and $\bm{D_x}(\bm{q}, \dot{\bm{q}}$), respectively. The system (\ref{eq:dynamic}) is converted from joint space to task space, which can be expressed as 
\begin{align}
\label{eq:convdyna}
\begin{bmatrix}
\dot{\bm{x}}\\
\ddot{\bm{x}}
\end{bmatrix}=\underbracket[0.5pt]{\begin{bmatrix}
\dot{\bm{x}}\\
(-\bm{M_x}^{-1})(\bm{C_x}\dot{\bm{x}}+\bm{g_x} + \bm{D_x})
\end{bmatrix}}_{f(\bm{x}, \dot{\bm{x}})}
+
\underbracket[0.5pt]{\begin{bmatrix}
0\\
-\bm{M_x}^{-1}
\end{bmatrix}}_{g(\bm{x})}\bm{F},
\end{align}
Therefore, AECBF can be applied to the robotic system (\ref{eq:convdyna}) to deal with the time-varying output constraints.

\subsection{Control Objectives}
The main objective of this paper is to design a novel control scheme for a robotic manipulator with the task space dynamic model inferred from the joint space dynamic model that has to track the pre-defined trajectory while avoiding collision with humans and obtain continuous output constraints. The performance of path tracking is improved by providing an RBFNN-based friction compensator. Additionally, a new optimization based safety-filter named AECBF for collision avoidance is presented to enable the robot to reach the closest position to the safe area by employing an adaptive ECBF coefficient. We take advantage of a machine learning algorithm toward this target to develop a mapping function between the generated force input and the minimum safe distance. We add these two modules into the original TVIBLF-ECBF controller step by step and design an ablation study to test their performance in an environment with static and dynamic obstacles.

Given the current human motion sequence $\bm{p_o}$ and the current position of the robot arm $\bm{x}(t)$, and assuming that: the dynamics of the manipulator are partially known, the objective is to determine a torque input $\bm{\tau}_{\rm act}$ to the robot, which enables the robot to track the desired trajectory $\bm{x_d}$ and avoid collision with the static and dynamic obstacles defined by $\bm{p_o}$.

\section{Radial Basis Function Neural Network for Uncertainty Estimation}
A radial basis neural network, commonly utilized to estimate uncertainties in model dynamics, is a neural network architecture consisting of three layers, i.e., an input layer, a hidden layer, and an output layer. In this work, an RBFNN is used to approximate the continuous function $\hat{\bm{D}}_{\bm{x}}(\bm{\chi})$, with a mapping of the form: 
\begin{align}
\label{eq:hatDx}
    \hat{\bm{D}}_{\bm{x}}(\bm{\chi}) = \hat{\bm{W}}^{\top}\bm{s}(\bm{\chi}),
\end{align}
Here, $\bm{\chi} = [\bm{F}, \bm{q}, \dot{\bm{q}}] \in \R^{2m+n}$ denotes the input vectors including the robot state and the Cartesian control force. The weights matrix of the RBFNN is $\hat{\bm{W}} = [\hat{\bm{w}}_1, \dots, \hat{\bm{w}}_{n}] \in \R^{l\times n}$, and $\bm{s}(\bm{\chi}) = [s_1(\bm{\chi}), s_2(\bm{\chi}), \dots, s_l(\bm{\chi})]^{\top} \in \mathbb{R}^l$ denotes the vector of hidden node outputs, where each hidden node $s_j(\bm{\chi})$ is a Gaussian basis function, 
\begin{align}
    s_j(\bm{\chi}) = \exp[\frac{-(\bm{\chi} - \bm{o}_j)^{\top}(\bm{\chi} - \bm{o}_j)}{\varsigma_j^2}], ~~j \in \N_{[1,l]},
\end{align}
Parameters $\bm{o}_j$ and $\varsigma_j$ define the centers and widths of the basis functions, respectively. There exists an optimal weight matrix $\bm{W}^{\star}$ , which yields
\begin{align}
   \bm{D}_{\bm{x}}(\bm{\chi}) = \bm{W}^{\star\top} \bm{s}(\bm{\chi}) + \bm{\epsilon}
\end{align}
where $\bm{\epsilon}\in \R^n$ is the approximation error. The optimal weight matrix $\bm{W}^{\star}$ is an artificial quantity for analytical purposes, which is defined as the value of $\hat{\bm{W}}$ that minimizes $|\bm{\epsilon}|$ for all $\bm{\chi} \in \bm{\Omega}_{\bm{\chi}} \subset \R^{2m+n}$
\begin{align}
    \bm{W}^{\star} = \argmin_{\hat{\bm{W}} \in \R^{l\times n}} \{ \sup_{\bm{\chi}\in \bm{\Omega}_{\chi}} |\bm{\epsilon}| \}.
\end{align}
The estimated system $\hat{f}(\bm{x}, \dot{\bm{x}})$ is obtained by substituting $\bm{D_x}$ in \eqref{eq:convdyna} with $\hat{\bm{D}}_{\bm{x}}$, as defined in \eqref{eq:hatDx}.

\section{Properties, Assumptions, and Lemmas}
\begin{myprop}
 The matrix $\bm{M}_x(\bm{q})$ is symmetric and positive definite.
\end{myprop}
\begin{myprop}
Under an appropriate definition of the matrix $\bm{C_x}$, the matrix $2\bm{C_x}(\bm{q}, \dot{\bm{q}}) - \dot{\bm{M}}_{\bm{x}}(\bm{q})$ is skew-symmetric, and as a consequence, for an arbitrary vector $\bm{z}$, it follows that
\begin{align}
\bm{z}^{\top}[2\bm{C_x}(\bm{q}, \dot{\bm{q}}) - \dot{\bm{M}}_{\bm{x}}(\bm{q})]\bm{z} = 0.
\end{align}
\end{myprop}
\begin{mydef}[Class $\mathcal{K}_{\infty}$-function \cite{grandia2020nonlinear}]
 A continuous function $\beta: [0, a)\rightarrow \R_{+}$, with $\beta = \infty$ and $\lim_{\bm{z}\rightarrow \infty}\beta(\bm{z}) = \infty$  is defined as a class $\mathcal{K}_{\infty}$-function.
 \end{mydef}

\begin{mydef}[Exponential Control Barrier Function (ECBF)]\label{def:ECBF}
    Consider the dynamic system \eqref{eq:cartesiandy} and the set $\mathcal{C}_1 = \{ \bm{x}\in \R^n| h(\bm{x}) \geq 0 \}$, where $h:\R^n\rightarrow \R$ has relative degree $r$. Then, $h(x)$ is an exponential control Barrier function (ECBF) if there exists $\bm{K_b}\in \R^r$ subject to
    \begin{align}
        inf_{\bm{u}\in \mathcal{U}}[L_f^r h(\bm{x}) + L_gL_f^{r-1}h(\bm{x})\bm{u} + \bm{K_b}\bm{\eta}(\bm{x})]\geq 0, \forall \bm{x}\in \mathcal{C}_1,
    \end{align}
    and $h(\bm{x}(t))\geq \bm{C_b}e^{\bm{A_b}t}\bm{\eta}(\bm{x}\geq 0)$, when $h(\bm{x}_0)\geq 0$, where $\bm{A_b}$ is a designed matrix on the choice of $\bm{K_b}$, and    
\end{mydef}
\begin{align}
        \bm{\eta}(\bm{x}) = \begin{bmatrix}
        h(\bm{x})\\
        \dot{h}(\bm{x})\\
        \vdots\\
        h^{(r-1)}(\bm{x})
\end{bmatrix}
        =
\begin{bmatrix}
          h(\bm{x})\\
        L_f h(\bm{x})\\
        \vdots\\
        L_f^{(r-1)} h(\bm{x})  
\end{bmatrix},
\end{align}
 
\begin{mylemma}[\cite{ren2010adaptive}] \label{lem:IBLF}
For any positive constants $k_{ci}$, $i\in \N_{[1,n]}$, let $\mathcal{C}_1:=\{ \bm{x}\in \R^n: - k_{a}<x_i<k_{b}\} \subset \R$ and $\mathcal{D}_1:=\R^l\times \mathcal{C}_1\subset \R^{n+l}$ be open sets. Given a system
\begin{align}
\label{eq:BLFproof}
    \dot{\bm{\eta}} = h(t, \bm{\eta}),
\end{align}
where $\bm{\eta} = [\bm{\omega}, \bm{x}]^{\top} \in \mathcal{D}_1$ and $h:\R^l\times \mathcal{D}_1 \rightarrow \R^{l+n}$ is piecewise continuous in $t$ and locally Lipschitz in $\bm{\eta}$ uniformly in $t$. Let $\mathcal{D}_{1i}:=\{x_i\in \R_{+}: |x_i(t)|<k_{ci}, i=\N_{[1,n]}, t\geq 0 \} \subset \R$.

Suppose that there exist functions $U: \R^l \rightarrow \R_{+}$ and $V_i: \mathcal{D}_{1i} \rightarrow \R_{+}$, $i=1,\dots,n$ continuously differentiable and positive definite in their respective domains, such that
\begin{align}
\label{eq:BLFineqchap2}
    &V_i \rightarrow \infty~~\text{as}~~|x_i|\rightarrow k_{ci},\\
    &\gamma_1(\Vert \bm{\omega} \Vert) \leq U(\bm{\omega}) \leq \gamma_2(\Vert \bm{\omega} \Vert),
\end{align}
where $\gamma_1$ and $\gamma_2$ are class $\mathcal{K}_{\infty}$ functions. 
Let $V(\bm{\eta}):=\sum_{i=1}^n V_{i}(x_i) + U(\bm{\omega})$, and $x_i(0) \in \mathcal{D}_1$. The state $\bm{x}$ will always be constrained in set $\mathcal{D}_1$, if the following inequality is satisfied
\begin{align}
\label{eq:IBLF}
    \dot{V} = \frac{\partial V}{\partial \bm{\eta}} \leq -\mu V + c
\end{align}
where $\mu$ and $c$ are positive constants.
\end{mylemma}

\begin{mylemma}[\cite{he2020admittance}]\label{lem:lyapunov}
To ensure the output of the system remains in the constrained task space, we introduce the IBLF candidate as
\begin{align}
    V = \sum_{i=1}^n V_i = \sum_{i=1}^n \int_0^{z_i} \frac{\sigma k_{ci}^2}{k_{ci}^2-(\sigma + x_i)^2}d\sigma,
\end{align}
where $z_i = x_i - x_{di}$, and $x_i$ is a continuously differentiable function satisfying $|x_i|<k_{ci}$, $i = \N_{[1,n]}$. It is known that $V$ is a continuously positive differentiable function over the set $\{|x_i| < k_{ci}\}$. As for $|x_i|<k_{ci}$, $i = \N_{[1,n]}$, there is
\begin{align}
    \frac{z_i^2}{2}\leq V_i \leq \frac{k_{ci}^2 z_i^2}{k_{ci}^2 - x_i^2}.
\end{align}
\end{mylemma} 

\begin{myassumption}
There exist positive constant 3D vectors $\underline{\bm{Y}} = [\underline{Y}_1, \underline{Y}_2, \underline{Y}_3]^{\top}$, $\bar{\bm{Y}} = [\bar{Y}_1, \bar{Y}_2, \bar{Y}_3]^{\top}$, such that, $\forall t\geq 0$, the desired trajectory of the $i^{\rm th}$ joint $\bm{x}_{d_i}(t)$ is bounded within $[-\underline{\bm{Y}},\bar{\bm{Y}}]$.
\end{myassumption}

\begin{mylemma} [\cite{he2020admittance}]\label{lem:RBFNN}
For the estimated RBFNN weight matrix $\hat{\bm{W}} = [\hat{w}_{ij}]$, there exists a compact set
\begin{align}
    \Omega_{w} = \{\hat{\bm{W}}| \Vert \hat{w}_{ij} \Vert\leq \bar{w}_{ij}\},
\end{align}
where $\hat{\bm{W}}(t) \in \Omega_{w}$, $~~\forall t \geq 0$ provided that $\hat{\bm{W}}(0) \in \Omega_{w}$.
\end{mylemma}


\section{TVIBLF For Continuously Output Constraint}
Denote error $\bm{z}_1 = \bm{x} - \bm{x_{d}}$, where $\bm{x}$ and $\bm{x_{d}}$ is the actual and desired position of the end-effector, respectively. In order to maintain the system output $\bm{x}$ within the time-varying constrained region, which is defined by $|x_{i}| < k_{di}$, $i\in \N_{[1,n]}$. We can consider the time-varying asymmetric IBLF as
\begin{align}
\label{eq:V1}
\begin{split}
    V_1 = &\sum_{i=1}^n \int_0^{z_{1i}}(p(z_{1i})\frac{\sigma k_{bi}^{2}}{k_{bi}^{2} - (\sigma+x_{di})^2}\\ &+(1-p(z_{1i}))\frac{\sigma k_{ai}^{2}}{k_{ai}^{2} - (\sigma+x_{di})^2})d\sigma,\\
\end{split}
\end{align}
where 
\begin{align}
    k_{ai}(t) &= k_{ci} + x_{di}(t),\\
    k_{bi}(t) &= k_{ci} - x_{di}(t),\\
    p(z_{1i}) &= 
    \begin{cases}
    1, \text{if $z_{1i} > 0$}\\
    0, \text{if $z_{1i} \leq 0$}
    \end{cases}.
\end{align}
Hence we can write $\dot{V}_1$ as:
\begin{align}
\label{eq:dV1}
\begin{split}
    \dot{V}_1 = &\sum_{i=1}^n (p \frac{z_{1i}k_{ai}^2}{k_{ai}^2-x_{1i}^2}\dot{z}_{1i} + (1-p) \frac{z_{1i}k_{ai}^2}{k_{ai}^2-x_{1i}^2}\dot{z}_{1i})\\
    &+  \sum_{i=1}^n (p\frac{\partial V_1}{\partial x_{di}}\dot{x}_{di} + (1-p)\frac{\partial V_1}{\partial x_{di}}\dot{x}_{di}).
    \end{split}
\end{align}
To simplify the equation (\ref{eq:dV1}), we define
\begin{align}
    k_{di} = \begin{cases}
    k_{bi}, ~~\text{if}~~z_{1i}>0\\
    k_{ai}, ~~\text{if}~~z_{1i} \leq 0
    \end{cases},
\end{align}
\label{eq:partialV1}
\begin{align}
    \frac{\partial V_1}{\partial x_{di}} = z_{1i}(\frac{k_{di}^2}{k_{di}^2 - x_{1i}^2} - \Phi_i).
\end{align}

According to L’Hôpital’s rule, we have
\begin{align}
    \lim_{z_{1i}\rightarrow 0} \Phi_i = \frac{k_{di}}{k_{di}^2-x_{di}^2}.
\end{align}

Define the tracking error $\bm{z}_2 = \bm{x_2} - \bm{\alpha}$, where $\bm{x_2}$ is the Cartesian velocity of the end-effector and $\bm{\alpha}$ is a virtual control variable. Its derivitive is given as $\dot{\bm{z}}_2 = \dot{\bm{x}}_2 - \dot{\bm{\alpha}}$. The virtual control variable $\alpha_i$ is designed as
\begin{align}
\label{eq:alpha}
    \alpha_i = -k_{zi}z_{1i} + \frac{(k_{di}^2 - x_{1i}^2)\dot{x}_{di}\Phi_i}{k_{di}^2},
\end{align}
where $k_{zi}$, $i = \N_{[1,n]}$ are positive constants. This allows $\dot{V}_1$ to be rewritten as 
\begin{align}
\label{eq:dotV1}
\begin{split}
    \dot{V}_1 = &\sum_{i=1}^n \frac{z_{1i}k_{di}^2}{k_{di}^2-x_{1i}^2}\dot{z}_{1i} + \sum_{i=1}^n z_{1i}(\frac{k_{di}^2}{k_{di}^2 - x_{1i}^2} - \Phi_i)\\
    =&\sum_{i=1}^n \frac{z_{1i}k_{di}^2}{k_{di}^2-x_{1i}^2}(z_{2i} + \alpha_i - \dot{x}_{di})\\
    &+ \sum_{i=1}^n z_{1i}(\frac{k_{di}^2}{k_{di}^2 - x_{1i}^2} - \Phi_i).
    \end{split}
\end{align}
To avoid the singularity for the second term in (\ref{eq:dV1}), $\Phi_i$ is defined as
\begin{align}
\begin{split}
    \Phi_i &= \int_0^1 \frac{k_{di}^2}{k_{di}^2 - (\sigma z_{1i} + x_{di})^2}d\sigma\\
    &= \frac{k_{di}}{z_{1i}}(\tanh^{-1}(\frac{z_{1i}+x_{di}}{k_{di}}) - \tanh^{-1}(\frac{x_{di}}{k_{di}}))\\
    &= \frac{k_{di}}{2z_{1i}}\ln \frac{(k_{di} + x_{1i})(k_{di} - x_{di})}{(k_{di} - x_{1i})(k_{di} + x_{di})}.
    \end{split}
\end{align}
Substituting (\ref{eq:alpha}) into (\ref{eq:dotV1}), we have

\begin{align}
    \dot{V}_1 = -\sum_{i=1}^n \frac{z_{1i}k_{zi} k_{di}^2}{k_{di}^2-x_{1i}^2} + \sum_{i=1}^n \frac{z_{1i}z_{2i}k_{di}^2}{k_{di}^2 - x_{1i}^2}.
\end{align}

Then, we design a TVIBLF candidate function as  
\begin{align}
\label{eq:V2}
   V_2 = V_1 + \frac{1}{2} \bm{z}_2^{\top} \bm{M_x} \bm{z}_2. 
\end{align}
Then, differentiating (\ref{eq:V2}) with respect to time yields
\begin{align}
\label{eq:dotV2}
\begin{split}
    \dot{V}_2 &= \dot{V}_1 + \frac{1}{2}\bm{z}_2^{\top}\dot{\bm{M}}_x \bm{z}_2 + \bm{z}_2^{\top}\bm{M_x}\dot{\bm{z}}_2 \\
     = &-\sum_{i=1}^n \frac{z_{1i}k_{zi}k_{di}^2}{k_{di}^2-x_{1i}^2} + \sum_{i=1}^n \frac{z_{1i}z_{2i}k_{di}^2}{k_{di}^2 - x_{1i}^2}\\
    & + \bm{z}_2^{\top}(\bm{F} - \bm{g_x} - \bm{C_x} (\bm{z}_2 + \bm{\alpha}) - \bm{D_x}\\
    &- \bm{M_x}\dot{\bm{\alpha}} +\frac{1}{2}\dot{\bm{M}}_x \bm{z}_2), 
\end{split}
\end{align}
where $\bm{D_x} = [D_{x1},\dots, D_{xn}] = \hat{\bm{W}}^{\top}\bm{s}(\bm{\chi}) - \bm{\epsilon}$. According to Property 2, we have
\begin{align}
\label{eq:dotV2true}
\begin{split}
    \dot{V}_2 & = -\sum_{i=1}^n \frac{z_{1i}k_{zi}k_{di}^2}{k_{di}^2-x_{1i}^2} +  \sum_{i=1}^n \frac{z_{1i}z_{2i}k_{di}^2}{k_{di}^2 - x_{1i}^2})\\
    & + \bm{z}_2^{\top}(\bm{F} - \bm{g_x} - \bm{C_x} \bm{\alpha} - \bm{M_x}\dot{\bm{\alpha}} - \hat{\bm{W}}^{\top}\bm{s}(\bm{\chi}) + \bm{\epsilon})
\end{split}
\end{align}

\section{Adaptive Neural Network Control}
According to Lemma \ref{lem:IBLF}, to satisfy the inequality function \eqref{eq:IBLF}. We design the control law as 
\begin{align}
\label{eq:f}
\begin{split}
    \bm{F} = \bm{C_x} \bm{\alpha}  + \bm{g_x} + \bm{M_x}\dot{\bm{\alpha}} + \hat{\bm{W}}^{\top}\bm{s}(\bm{\chi}) - \bm{K}_b \bm{z}_2 - \bm{\eta}
     \end{split}
\end{align}
where $\bm{\eta} = [\eta_1, \eta_2, \dots, \eta_n]$. $\bm{K_b}$ is an $n$-dimensional positive definite matrix and $\eta_i$ is expressed as
\begin{align}
\eta_i =  -\frac{k_{zi}z_{1i} k_{di}^2}{k_{di}^2 - x_{1i}^2}, i \in \N_{[1,n]} 
\end{align}

\noindent\textbf{Proof:}
Substituting (\ref{eq:alpha}) and (\ref{eq:f}) into (\ref{eq:dotV2true}), we have 
\begin{align}
    \dot{V}_2 = -\sum_{i=1}^n \frac{k_{zi}z_{1i}^2k_{di}^2}{k_{di}^2} - \bm{z}_2^{\top}\bm{K_b} \bm{z}_2 + \bm{z}_2^{\top}\bm{\epsilon}.
\end{align}
According to Young's inequality
\begin{align}
    \bm{z}_2^{\top}\bm{\epsilon}\leq \frac{1}{2c_1^2}\bm{z}_2^{\top}\bm{z}_2 + \frac{c_1^2}{2}\Vert \Bar{\bm{\epsilon}} \Vert^2
\end{align}
Considering Lemma \ref{lem:IBLF}, we can get
\begin{align}
\label{eq:ineqdV2proof}
\begin{split}
\dot{V}_2 &\leq -\sum_{i=1}^n\int_{0}^{z_{1i}}\frac{\sigma k_{zi} k_{di}^2}{k_{di}^2 - (\sigma + x_{di})^2}d\sigma\\
&- \bm{z}_2^{\top}(\bm{K_b} - \frac{1}{2c_1^2}\bm{I}) \bm{z}_2+ \frac{c_1^2}{2}\Vert \Bar{\bm{\epsilon}} \Vert^2\\
&\leq -\mu V_2 +c_2,
\end{split}
\end{align}
where $\bm{I}$ is an identity matrix and $\mu$ and $c_2$ are small positive constants defined as
\begin{align}
\begin{split}
    &\mu = \min(\min_{i=1,\dots,n} (k_{zi}), \frac{2\lambda_{min}(\bm{K_b} - \frac{1}{2c_1^2}\bm{I})}{\lambda_{min}(\bm{M_x})}),\\
    &c_2 = \frac{c_1^2}{2}\Vert \Bar{\bm{\epsilon}} \Vert^2
    \end{split}
\end{align}
where $\mu>0$ and the $\lambda_{min}$ function gives the minimum eigenvalue of the matrix. It is obvious that $V_2$ will converge to zero, and the task space position $\bm{x_1}$ can remain in the defined constrained space.

The adaptive law is given as follows:
\begin{align}
\label{eq:adaptlaw}
    \dot{\hat{\bm{W}}} = \bm{s}(\bm{\chi})\bm{z}_2^{\top} - \rho\hat{\bm{W}}, 
\end{align}
where $\rho > 0$ is a small constant. 

\noindent\textbf{Proof:}
As $\tilde{\bm{W}} = \bm{W}^{\star}- \hat{\bm{W}}$ denotes weights error, we have
\begin{align}
\begin{split}
    V_3 &= V_2 + \frac{1}{2}\sum_{i=1}^n\tilde{\bm{w}}_i^{\top}\tilde{\bm{w}}_i.
    \end{split}
\end{align}
 Then, differentiating $V_3$ yields
 \begin{align}
 \label{eq:dV3proof}
     \dot{V}_3 =\dot{V}_2 - \sum_{i=1}^n\tilde{\bm{w}}_i^{\top}\dot{\hat{\bm{w}}}_i.
 \end{align}
Substituting (\ref{eq:ineqdV2proof}) and (\ref{eq:adaptlaw}) into (\ref{eq:dV3proof}) yields
\begin{align}
\label{eq:dotV3proof}
\begin{split}
    \dot{V}_3 &= -\sum_i^n (\frac{z_{1i}k_{di}^2}{k_{di}^2-x_{1i}^2} + \frac{z_{1i}z_{2i}k_{di}^2}{k_{di}^2 - x_{1i}^2}\\
    & + \bm{z}_2^{\top}(\bm{F} - \bm{g_x} - \bm{C_x} \bm{\alpha} - \bm{M_x}\dot{\bm{\alpha}} - \bm{D_x})\\
    &\leq -\sum_{i=1}^n \frac{k_{zi}z_{1i}^2k_{di}^2}{k_{di}^2} - \bm{z}_2^{\top}\bm{K}_b\bm{z}_2\\
    &+\bm{z}_2^{\top}(\hat{\bm{W}}^{\top}\bm{s}(\bm{\chi}) - \bm{W}^{\star\top}\bm{s}(\bm{\chi}) + \bm{\epsilon})\\
    &-\sum_{i=1}^n\tilde{\bm{w}}_i^{\top}(\bm{s}(\bm{\chi})z_{2i} - \rho\hat{\bm{w}}_i),
\end{split}
\end{align}
where $\bm{D_x} = \bm{W}^{\star}\bm{s}(\bm{\chi})$. According to Young's inequality
\begin{align}
\begin{split}
    \rho\tilde{\bm{w}}_i^{\top}\hat{\bm{w}}_i &= -\rho\tilde{\bm{w}}_i^{\top}\tilde{\bm{w}}_i + \rho\tilde{\bm{w}}_i^{\top}\bm{w}_i^{\star}\\
    &\leq -\frac{\rho}{2}\tilde{\bm{w}}_i^{\top}\tilde{\bm{w}}_i + \frac{\rho}{2} \Vert\bm{w}_i^{\star}\Vert,
    \end{split}
\end{align}
where $\bar{\bm{\epsilon}}$ represents the upper boundary of $\bm{\epsilon}$ and $c_1$ is a small positive constant. Then $\dot{V}_3$ is expressed as
\begin{align}
\begin{split}
    \dot{V}_3 \leq &-\sum_{i=1}^n k_{zi} \int_{0}^{z_{1i}}\frac{\sigma k_{di}^2}{k_{di}^2 - (\sigma + x_{di})^2}d\sigma\\
    &- \bm{z}_2^{\top}(\bm{K}_b-\frac{c_1^2}{2}\bm{I})\bm{z}_2 - \frac{\rho}{2}\sum_{i=1}^n \tilde{\bm{w}}_i^{\top}\tilde{\bm{w}}_i\\
    & + \frac{c_1^2}{2}\Vert \bar{\bm{\epsilon}}\Vert^2 +\frac{\rho}{2}\sum_{i=1}^n \Vert \bm{w}_i^{\star} \Vert^2\\
    &= -\mu_1 V_3 + c_3,
\end{split}
\label{eq:dotV3ineq}
\end{align}
where positive constants $\mu_1$ and $c_3$ are expressed as
\begin{align}
    \begin{split}
    &\mu_1 = \min(\min_{i=1,\dots,n} (k_{zi}), \frac{2\lambda_{min}(\bm{K_b} - \frac{1}{2c_1^2}\bm{I})}{\lambda_{min}(\bm{M_x})}, \rho),\\
    &c_3 =  \frac{c_1^2}{2}\Vert \bar{\bm{\epsilon}}\Vert^2 +\frac{\rho}{2}\sum_{i=1}^n \Vert \bm{w}_i^{\star} \Vert^2.
    \end{split}
\end{align}
\section{AECBF-based Safety Filter}
In this section, an ECBF-based safety filter is introduced to prevent collisions with the human operator by constraining the trajectory generated by NMPC to be in a safe zone. In order to achieve asymptotic stability, a CLF constraint is also required to be satisfied. 
Denote $\bm{p}_{oi}(t) \in \mathbb{R}^3$ as the center of the $i^{\rm th}$ obstacle at time $t$. To ensure the robot does not collide with the moving human, the Cartesian position of the robot's $j^{\rm th}$ joint $\bm{x}_j \in \R^3$, $~~j \in \N_{[1, m]}$ are under the following constraints
\begin{align}
\label{eq:safedistance}
\begin{split}
    \Vert \bm{x}_j(t) - \bm{p}_{oi}(t) \Vert_2 &> d_{safe}\\
    &,j \in \N_{[1, m]}, i\in \N_{[1,N_o]},  t \geq 0,
    \end{split}
\end{align}
where $N_o$ is the number of obstacles and $d_{safe}=r_i+d_m$ is the minimum allowable distance. Here  $r_i$ is the radius of the $i^{\rm th}$ obstacle and $d_m$ is a user specified safety margin.
Denoting the distance between the $i^{\rm th}$ obstacle and the $j^{th}$ robot's joint as $\bm{Z} = [\bm{\zeta}_1,\dots,\bm{\zeta}_{N_o}]$, where $\bm{\zeta}_i = \bm{x}_{j}(t) - \bm{p}_{oi}(t)$. We define a candidate ECBF of the $j^{\rm th}$ joint of the robot $h(t, \bm{x_j}, \bm{p}_{oi})$ as
\begin{align}
\label{eq:hchap5}
h(t, \bm{x}_{j}, \bm{p}_{oi}) = \Vert \bm{\zeta}_i \Vert^2 - d_{safe}^2,
\end{align}
The first-order lie derivative is given by
\begin{align}
    L_{\hat{f}}h(t, \bm{x}_{j}, \bm{p}_{oi}) = 2\bm{\zeta}_{i}^{\top}(\dot{\bm{x}}_{j}-\dot{\bm{p}}_{oi}),
\end{align}
and the second-order lie derivative is given by
\begin{align}
    L_{\hat{f}}^2 h(t, \bm{x}_{j}, \bm{p}_{oi}) =  2\bm{\zeta}_{i}^{\top}(\ddot{\bm{x}}_{j}-\ddot{\bm{p}}_{oi}) + 2\Vert \dot{\bm{x}}_{j} -  \dot{\bm{p}}_{oi} \Vert^2.
\end{align}
Thus, according to Definition \ref{def:ECBF}, the ECBF constraint is designed as
\begin{align}
\label{eq:lie2constraint}
\begin{split}
     L_{\hat{f}}^2 h(t,\bm{x}_j, \bm{p}_{oi}) &+ k_1 h(t,\bm{x}_j, \bm{p}_{oi})\\
    &+ k_2 L_{\hat{f}} h(t, \bm{x}_j, \bm{p}_{oi}) \geq 0,\\
    &j\in \N_{[1,m]}, i\in \N_{[1,N_o]},
\end{split}
\end{align}
where $h(t,\bm{x}_j, \bm{p}_{oi})$ is as defined in (\ref{eq:hchap5}) and $L_{\hat{f}}$ is the lie derivative of $\hat{f}(\bm{x})$.

To modify the applied forces $\bm{F}$ based on the safety constraint \eqref{eq:lie2constraint} and  simultaneously achieve asymptotic stability, a quadratic programming (QP)-based safety filter is employed. Considering the position constraints with relative degree 2, the ECBF optimization problem can be formulated by minimizing the normed difference between $\bm{F}$ and the actual input $\bm{u}_{\rm act}$, that is,

\begin{align}
\label{eq:ecbfQP}
    \bm{u}_{\mathrm{act}}= \argmin_{\bm{u}_\mathrm{act} \in \mathcal{U}} \Vert \bm{F} - \bm{u}_{\mathrm{act}} \Vert^2,
\end{align}
subject to
\begin{subequations}
\begin{align}
\label{eq:ecbfconstraint}
    -2\bm{\zeta}_{i}^{\top}\bm{M_x}^{-1}\bm{u}_{\rm act, \textit{j}} \leq& 
    2\bm{\zeta}_{i}^{\top}\ddot{\bm{x}}_{j}+ k_1 h(t,\bm{x}_j, \bm{p}_{oi})\\
    +k_2 L_{\hat{f}} h(t, \bm{x}_j, \bm{p}_{oi})&, j\in \N_{[1,m]},i\in \N_{[1,N_o]}\notag\\
    V_4(\bm{z}_2, \bm{u}_{\rm act}, t) &\leq \mu_2 V_4 + c_6,
\end{align}
\end{subequations}
where $\mu_2$ and $c_6$ are positive constants. The Lyapunov candidate $V_4$ is defined as
\begin{align}
\label{eq:V4}
    V_4 = \frac{1}{2} \bm{z}_2^{\top} \bm{M_x} \bm{z}_2 + \frac{1}{2}\sum_{i=1}^n\tilde{\bm{w}}_i^{\top}\tilde{\bm{w}}_i,
\end{align}

\subsection{Multi-layer Perceptron to Build the Penalty Term}
\begin{figure}
    \centering
    \includegraphics[width=0.48\textwidth]{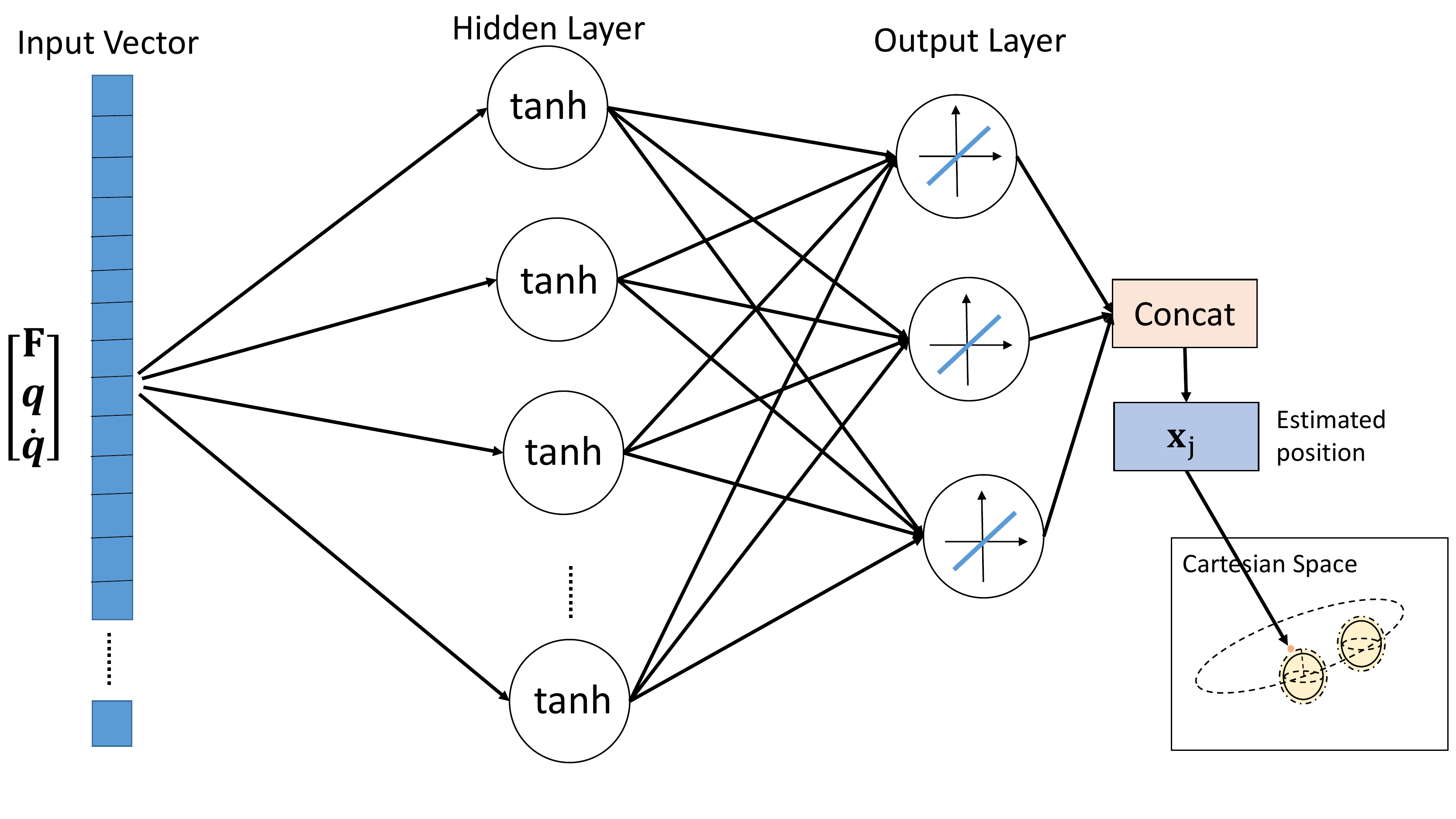}
    \caption{Estimating the Cartesian position of the robot's $j^{\text{th}}$ joint based on the control input $\bm{u}_{\rm act}$, the joint position and velocity using a multi-layer perceptron (MLP).}
    \label{fig:FNNsetup}
\end{figure}

In order to solve the quadratic penalty problem of AECBF, we address the control in the continuous domain by designing a sufficiently small sampling period. An AECBF problem is designed to achieve optimal trajectory for collision avoidance by adding a penalty term reflecting the distance to obstacles and how to achieve the minimum movement to execute collision avoidance. Towards this target, we build a function $\Phi(\bm{x}|\bm{u},\bm{q},\dot{\bm{q}})$ with using a Multilayer Perceptron (MLP) neural network to approximate the Cartesian position of the robot. The architecture of the MLP is shown in Fig. \ref{fig:FNNsetup}.
At each sampling time, the quadratic penalty problem is described as 
\begin{align}
\label{eq:AECBFpenalty}
\begin{split}
   &\argmin_{\bm{u}_{\text{act},j}^k \in \mathcal{U}} \Vert \bm{F}^k - \bm{u}_{\text{act},j}^k \Vert^2\\
   &+\sum_{i=1}^{N_o}\Vert \Phi(\bm{x}_{j}^k|\bm{u}_{\rm act,\textit{j}}^k, \bm{q}_j^k, \dot{\bm{q}}_j^k) - \bm{p}_{\rm oi}^k - d_{\rm safe}\bm{\Gamma}\Vert^2,\\
   &j\in \N_{[1,m]}, i\in \N_{[1,N_o]},
\end{split}
\end{align}
where $\bm{\Gamma} = [1,1,1]$. Therefore, the solution to the quadratic penalty problem enables the robot to reach the closest position to the boundary of obstacles, and finally finish the collision avoidance task with the minimum movement.
Then, the input to the robot $\bm{\tau}^k$ at each sampling time is transformed from $\bm{u}_{\rm act}^k$ as $\bm{\tau}^k = \bm{J}^{\top}(\bm{q}^k)\bm{u}_{\rm act}^k$. The details of the NN-TVIBLF-AECBF algorithm are shown in Algorithm \ref{alg:chapter5}. 

\subsection{Stability Analysis}
According to \eqref{eq:V4} with $\bm{F}$ replaced by the modified control force $\bm{F} - \bm{u}_{\rm act}$, the time derivative of $V_4$ yields 
\begin{align}
\label{eq:dotV4}
\begin{split}
    \dot{V}_4 =  \bm{z}_2^{\top}(\bm{F} - \bm{u}_{\rm act} - \bm{g_x} - \bm{C_x} \bm{\alpha} &- \bm{D_x} - \bm{M_x}\dot{\bm{\alpha}})\\
    &- \sum_{i=1}^n \tilde{\bm{w}}_i^{\top}\dot{\hat{\bm{w}}}_i,
\end{split}
\end{align}
then substituting \eqref{eq:f} into \eqref{eq:dotV4} we have
\begin{align}
\begin{split}
    \dot{V}_4 =  \bm{z}_2^{\top}( - \bm{u}_{\rm act} + \bm{K_b} \bm{z}_2 &- \hat{\bm{W}}\bm{s}(\bm{\chi}) + \bm{\epsilon})\\
    &- \sum_{i=1}^n \tilde{\bm{w}}_i^{\top}\dot{\hat{\bm{w}}}_i.
    \end{split}
\end{align}
According to Young's inequality
\begin{align}
\label{eq:uactineq}
    \bm{z}_2^{\top}\bm{u}_{\rm act} \leq \frac{1}{2c_4^2}\bm{z}_2^{\top}\bm{z}_2 + \frac{c_4^2}{2} \Vert \bm{u}_{\rm act}\Vert^2,
\end{align}
where $c_4$ is a small positive constant. According to \eqref{eq:uactineq} and \eqref{eq:dotV3ineq}, it follows that
\begin{align}
\begin{split}
    \dot{V}_4 \leq &- \bm{z}_2^{\top}(\bm{K_b} - (\frac{1}{2c_1^2}+\frac{1}{2c_4^2})\bm{I})\bm{z}_2 - \frac{\rho}{2}\sum_{i=1}^n \tilde{\bm{w}}_i^{\top}\tilde{\bm{w}}_i\\
    &+ \frac{c_1^2}{2}\Vert \bar{\bm{\epsilon}}\Vert^2 + \frac{c_3^2}{2}\Vert \bm{u}_{\rm act}\Vert^2 + \frac{\rho}{2}\sum_{i=1}^n \Vert \bm{w}_i^{\star} \Vert^2\\
    &=-\mu_2 V_4 + c_5,
    \end{split}
\end{align}
where
\begin{align}
    \begin{split}
    &\mu_2 = \min(\frac{2\lambda_{min}(\bm{K_b} - (\frac{1}{2c_1^2}+\frac{1}{2c_4^2})\bm{I})}{\lambda_{min}(\bm{M_x})}, \rho),\\
    &c_5 = \frac{c_1^2}{2}\Vert \bar{\bm{\epsilon}}\Vert^2 + \frac{c_4^2}{2}\Vert \bm{u}_{\rm act}\Vert^2 + \frac{\rho}{2}\sum_{i=1}^n \Vert \bm{w}_i^{\star} \Vert^2.
    \end{split}
\end{align}

\begin{algorithm}
\DontPrintSemicolon
  
  \KwInput{$\bm{x}_d \in \R^3$ (desired trajectory of the end-effector), $\bm{x}_0 \in \R^3$ (Initial position of the end-effector).}
  \KwOutput{The actual input to the robot $\bm{\tau}_{\rm act}$.}
  \KwData{The positions of obstacles $\bm{x}_{o,k}$ detected from sensors.}
  Create a random initial weight $\hat{\bm{W}}_0$ for RBFNN.\;
  \For{$t = 1,\dots, T$}{
  $\bm{z_1} = \bm{x} - \bm{x}_{d}$.\;
  Compute the virtual input $\bm{\alpha}$.\;
  $\bm{z}_2 = \dot{\bm{x}} - \bm{\alpha}$.\;
  Update the weight of the RBFNN (using \eqref{eq:adaptlaw}) and estimate the friction matrix $\hat{\bm{D}}_{\bm{x}}$.\;
  Compute the nominal control force $\bm{F}$.\;
  Estimate the actual dynamics by\; 
  $\hat{f}(\bm{x}(t)) = \bm{M}_{\bm{x}}^{-1}(\bm{F} - \bm{C}_{\bm{x}} \dot{\bm{x}}(t) -\bm{g}_{x} - \hat{\bm{D}}_{\bm{x}} + \bm{\epsilon})$.\;
  \If{$\min \Vert \bm{x}_j(t) - \bm{p}_{oi}(t) \Vert \leq d_{\rm safe}$}{
   Solve the AECBF quadratic penalty program problem.\;
   Compute the input to the robot $\bm{\tau}_{\rm act}(t) = \bm{J}^{\top}(\bm{q}(t))\bm{u}_{\rm act}(t)$.\;
  }
   \Else{
   The input to the robot is $\bm{\tau}_{\rm act}(t) = \bm{J}^{\top}\bm{F}(t)$.\;
  }
}
\caption{Algorithm for NN-TVIBLF-AECBF}
\label{alg:chapter5}
\end{algorithm}
\section{Experiments and results}
This section employs a 7-DOF KUKA LBR iiwa robot to test our proposed approach. The simulation experiments are performed in Simulink. The experiments are designed to verify that the robot can perform the pre-defined path-tracking task and guarantee the safety of the human operator by avoiding collision with the obstacle on the desired trajectory. In our simulation studies, we arbitrarily define the safety margin as $d_m = 7$ cm, given the dimensions of the robot and obstacles considered.

\begin{figure}
    \centering
    \includegraphics[width=0.45\textwidth]{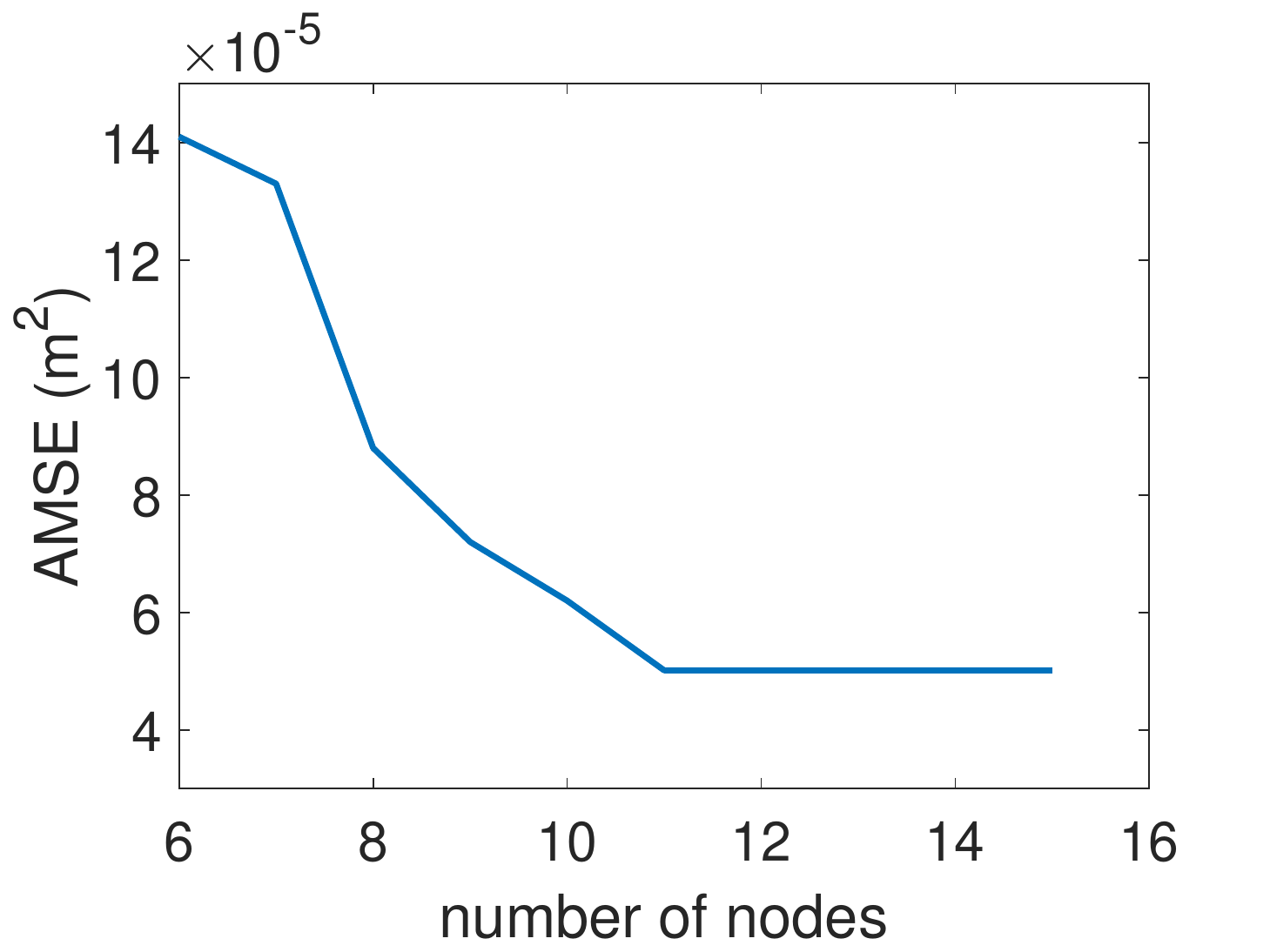}
    \caption{The relationship between the average mean sqaure error (AMSE) and the number of nodes.}
    \label{fig:node}
\end{figure}

In the control design part, we design an adaptive IBLF-ECBF controller with output constraints and analyze the stability of the system by the Lyapunov method. Initially, the end-effector starts from a initial position $\bm{x}_0 = [-0.13(m), -0.4(m), 0.74(m)]$. The constraints on state $\bm{k_c}$ are set as $[0.6,0.95,1.2]$. Control parameters are defined as $c_2=10$, $\bm{k}_{z} = [17.5,15,22.2]$ and $\bm{K_b} = diag[11.4,12,4.5]$. To determine the optimum RBFNN size we compared the average mean square error between the actual trajectory and the desired trajectory in path tracking for RBFNNs with different numbers of neurons, as shown in Fig. \ref{fig:node}. This shows that an RBFNN with 11 neurons achieves the minimum AMSE. The centers of the RBFNN nodes are distributed within the motion range of the joint space which is [-3, 3]. The settings of the centers ensures the RBFNN traverses the whole joint space to provide good uncertainty estimation. The initial weights of the RBFNN are generated randomly. The parameter of the RBFNN adaptive law is $\rho = 0.4$. 

\begin{table}[htbp]
  \centering
  \caption{Total runtime in each iteration, the AMSE, and maximum MSE between the predicted task-space state and the desired state in path tracking under different hidden sizes.}
  \small\addtolength{\tabcolsep}{-5pt}
  \resizebox{\linewidth}{!}{%
    \begin{tabular}{|c|c|c|c|c|c|}
    \hline
          & size=32 & size=64 & size=128 & size=256 & size=512 \\
    \hline
    run time (ms) & 8.3  & 9.7  & 10.3  & 13.7  & 15.6 \\
    \hline
    AMSE ($m^2$) & 1.60E-04 & 1.22E-04 & 1.06E-04 & 9.50E-05 & 9.49E-05 \\
    \hline
    MSE\_max ($m^2$) & 3.66E-04 & 2.98E-04 & 2.57E-04 & 2.44E-04 & 2.42E-04 \\
    \hline
    \end{tabular}}%
  \label{tab:MLPsize}%
\end{table}%
The MLP is trained using an epoch size of 200 with a hidden layer of 64 tanh neurons using high-quality training data collected from a  KUKA LBR simulation. To determine the appropriate number of neurons we compared the total runtime in each 10 ms control sample period, the AMSE, and the maximum MSE between the predicted task-space state and the desired state in path tracking for MLPs with different hidden layer sizes (i.e. 32, 64, 128, 256 and 512 neurons), as shown in Table. \ref{tab:MLPsize}. In each control sample, the computation of the TVIBLF-based control force generation, AECBF-based safety filter, RBFNN-based uncertainty estimation and the robot execution are included.  Although the larger networks provide more accurate state estimation, they result in a longer run time in each iteration, and exceed the 10 ms computation time limit for real-time implementation. Thus, an MLP with 64 neurons was selected as the optimum trade-off between computation time and accuracy.
The input to the MLP is a vector consisting of the current control force $\bm{F} \in \R^{7\times 3}$, the angular position $\bm{q}\in \R^7$ and the angular velocity $\dot{\bm{q}}\in \R^7$, while the output is the robot's Cartesian position $\bm{x} \in \R^{7\times 3}$. The Levenberg–Marquardt algorithm is used to train the model.

\begin{figure}
    \centering
     \subfloat[]{
    \includegraphics[width=0.45\textwidth]{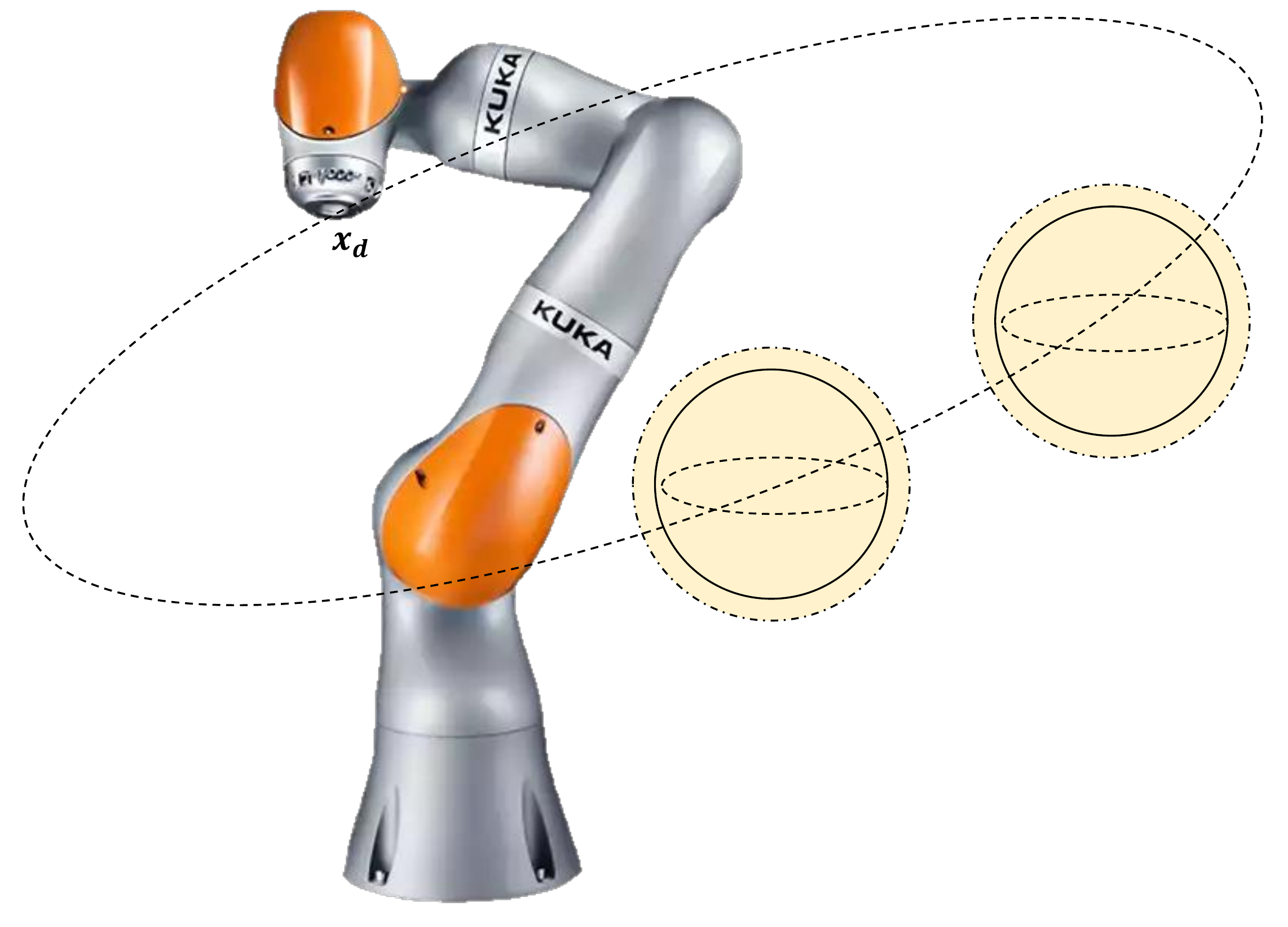}}\hfil
     \subfloat[]{
     \includegraphics[width=0.45\textwidth]{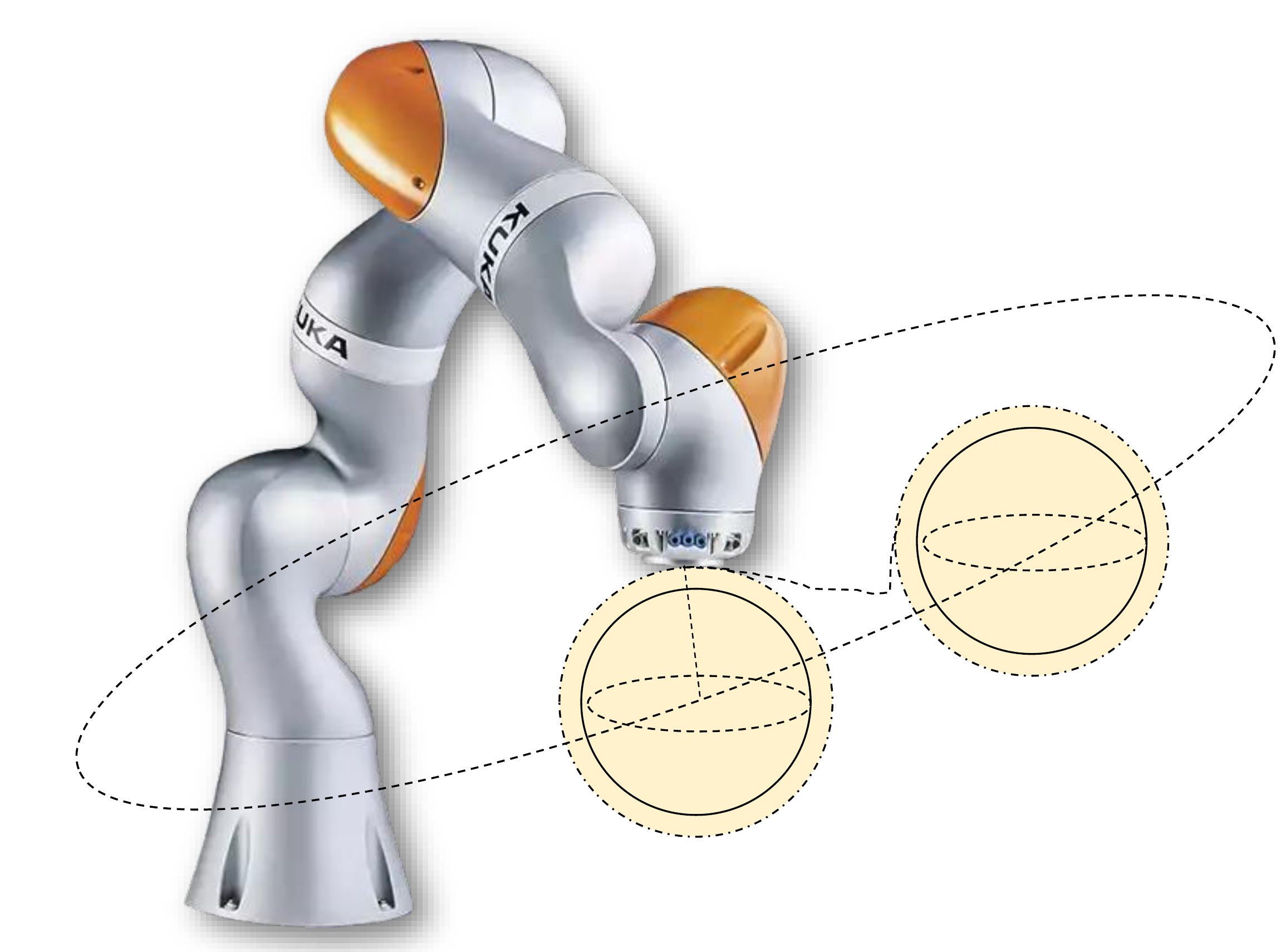}}\hfil
    \caption{(a) Simulation of a path tracking task. (b) Simulation of a collision avoidance task.}
    \label{fig:ptcvvisualization}
\end{figure}

The desired end-effector trajectory in the task space is 
\begin{align}
\begin{split}
    x_{d_x}(t) = (0.2sin(2t)-0.1)(m),\\
    x_{d_y}(t) = (0.2cos(2t)-0.6)(m),\\
    x_{d_z}(t) = (0.2sin(2t)+0.75)(m).\\
    \end{split}
\end{align}
We evaluate the performance of the NN-TVIBLF-AECBF controller by comparing it with the baseline TVIBLF-ECBF and NN-TVIBLF-ECBF controllers. We first add the friction compensator to the baseline, which is denoted by NN-TVIBLF-ECBF. Then we replace the ECBF module with AECBF, namely NN-TVIBLF-AECBF, and we evaluate the performance of a collision avoidance task from the perspective of the total movement, total rotation, and maximum displacement from the desired trajectory. Fig. \ref{fig:ptcvvisualization} (a) and (b) plot the robot's behavior under a path-tracking task and a collision avoidance task, respectively. The dashed line around the 3D obstacle represents the safe area.

\subsection{Case1: Collision Avoidance with Static Obstacles}
In scenario 1, an ECBF-based safety filter is triggered when the minimum distance between the robot and obstacles reaches the required safe distance under the supervision of the perception system. 

\begin{figure}
    \centering
    \includegraphics[width=0.5\textwidth]{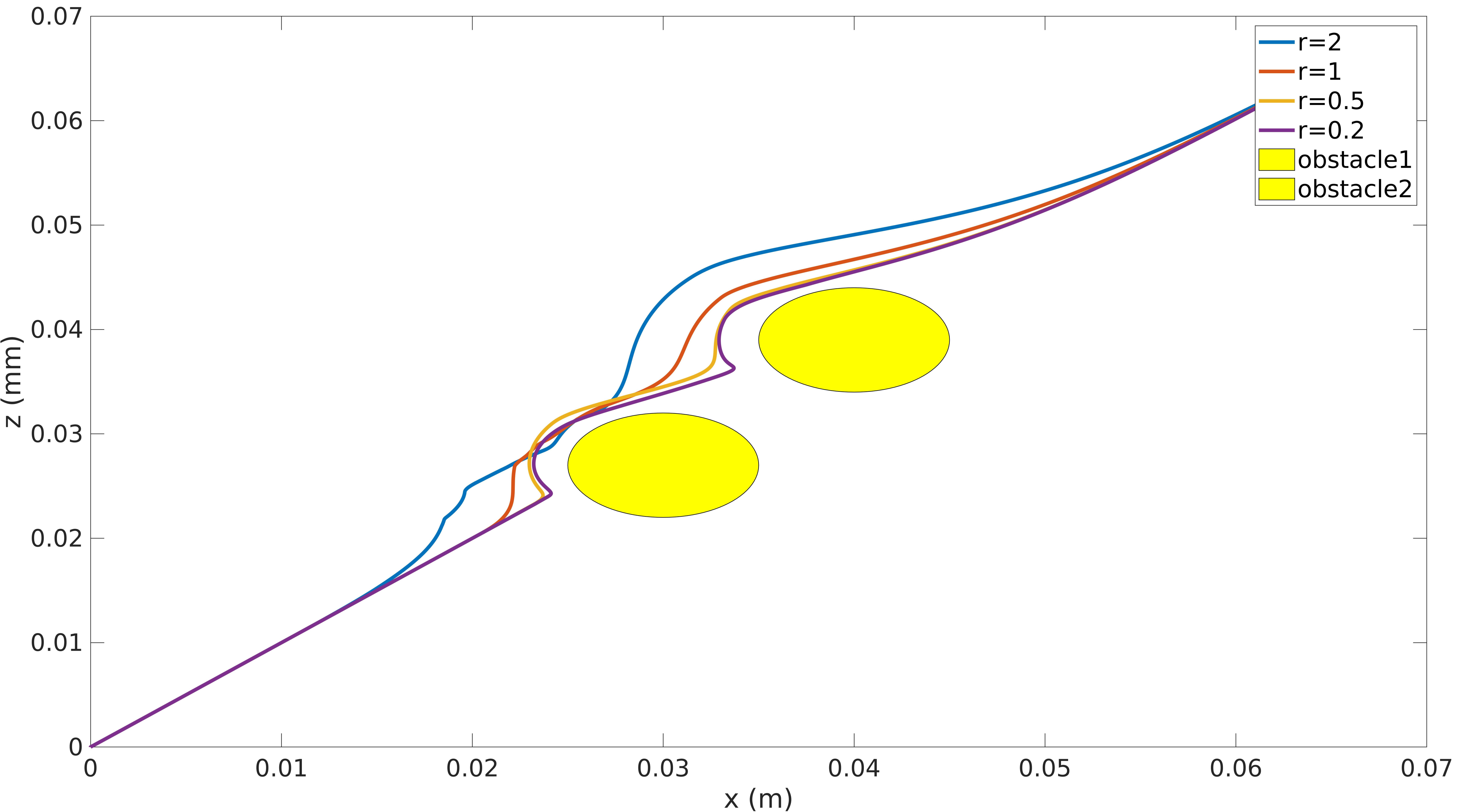}
    \caption{Simulations of the TVIBLF-ECBF for collision avoidance under different rate $r$.}
    \label{fig:ECBFrate}
\end{figure}

 In order to decide the proper combination of constraint weighting factors $k_1$ and $k_2$ in (\ref{eq:ecbfconstraint}), we define the ratio $r = \frac{k_1}{k_2}$ and plot the trajectories generated by ECBF for different values of $r$, as shown in Fig. \ref{fig:ECBFrate}. Based on this comparison, we select $r = 0.2$ which is the value that yields the collision avoidance trajectory with the minimum deviation from the obstacle-free path.

Table. \ref{tab:totalmove} records the total joint revolution of the robot, which is defined as the sum of the integral of the task space velocity of the end-effector and the joint velocities over time (summed over all joints). It shows that both the friction compensator and AECBF contribute to generating a shorter trajectory in Cartesian space with less joint rotation.

\begin{table}[htbp]
  \centering
  \caption{Total movement and rotation to execute the designed path tracking and static obstacle avoidance task on three models.}
  \small\addtolength{\tabcolsep}{-5pt}
  \resizebox{\linewidth}{!}{%
    \begin{tabular}{|c|c|c|}
    \hline
          & distance (m) & rotation (degrees) \\
    \hline
    TVIBLF-ECBF & 4.36  & 949.1 \\
    \hline
    NN-TVIBLF-ECBF & 4.11  & 910.2 \\
    \hline
    NN-TVIBLF-AECBF & $\bm{3.92}$  & $\bm{892.8}$ \\
    \hline
    \end{tabular}}%
  \label{tab:totalmove}%
\end{table}%



\begin{figure}
     \centering
    \subfloat[]{\includegraphics[width=0.9\linewidth]{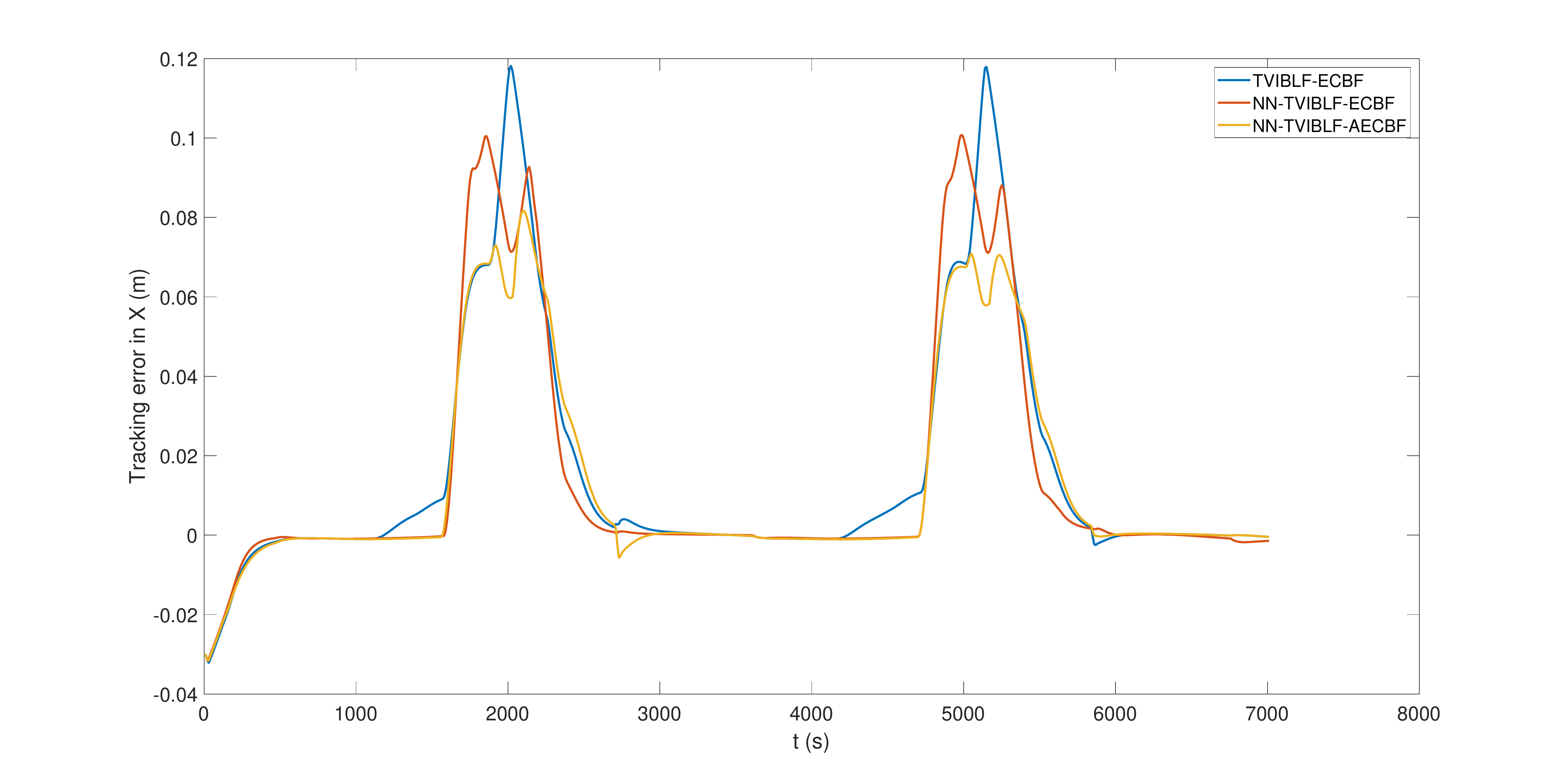}}\hfil
    \label{ob1}
    \subfloat[]{\includegraphics[width=0.9\linewidth]{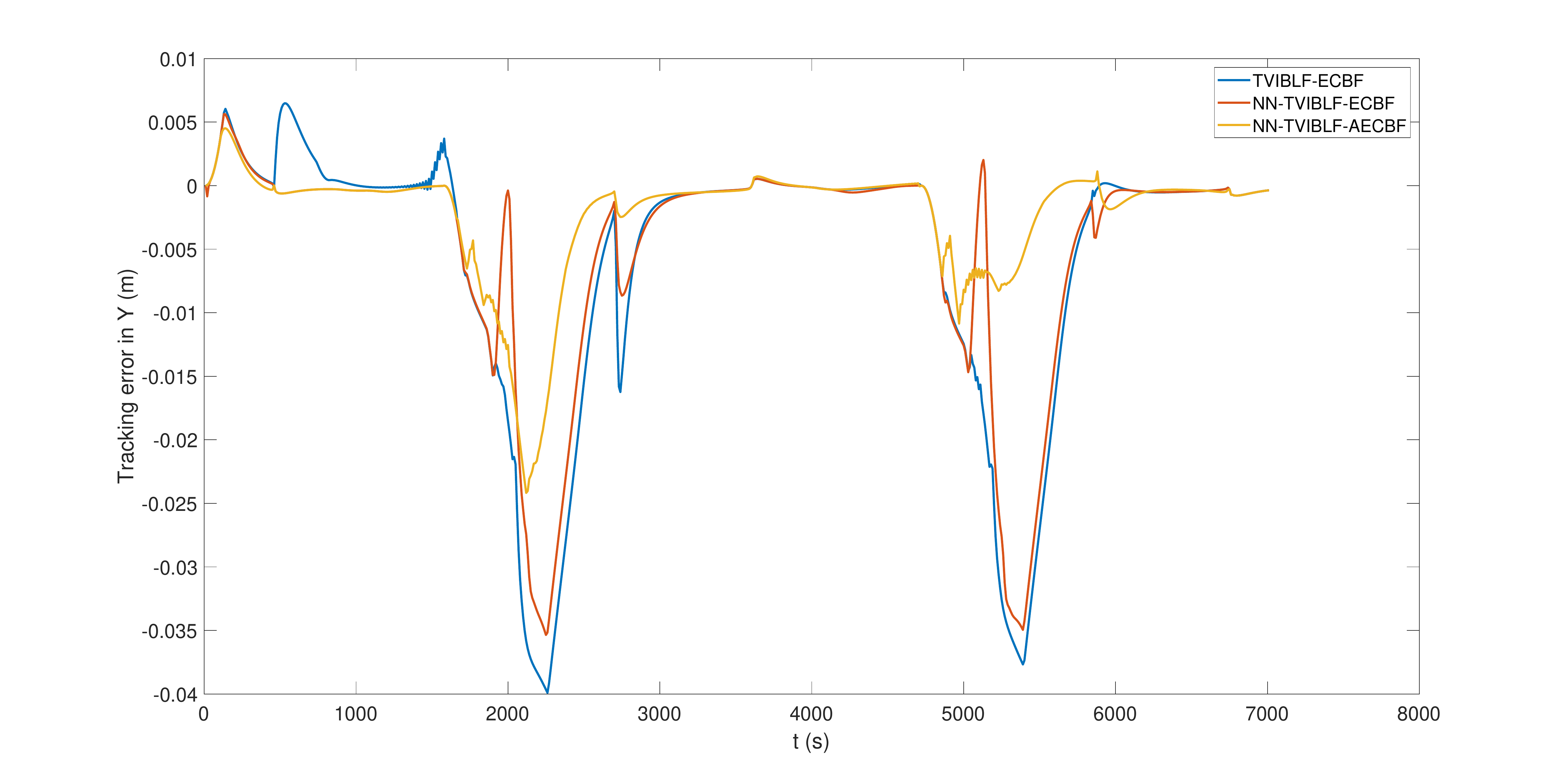}}\hfil
    \label{ob2}
     \subfloat[]{\includegraphics[width=0.9\linewidth]{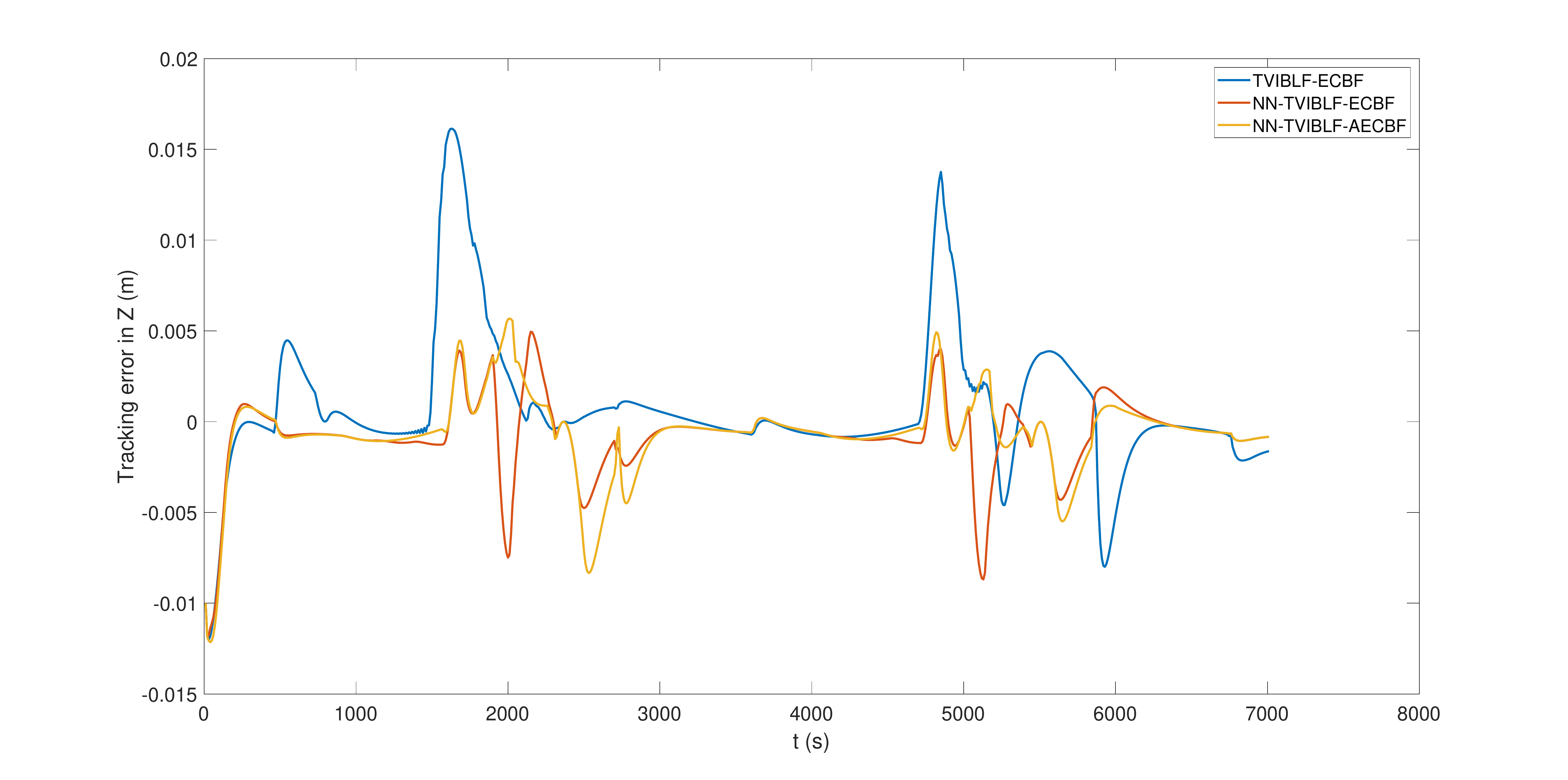}} 
     \caption{Comparison of the tracking errors along the X, Y, and Z axes.}
     \label{fig:trackerrorstatic}
 \end{figure}

	
Table \ref{tab:eAstatic} and Table \ref{tab:eBstatic} measure the maximum error in path tracking and collision avoidance, respectively. We find the performance of the TVIBLF-ECBF is improved when using the friction model. After adding an RBFNN-based friction compensator, a 7.5\% reduction in the euclidean distance path tracking error is obtained when compared with TVIBLF-ECBF. The best performance is obtained by the NN-TVIBLF-AECBF model, with an 85\% reduction in error for path tracking and 24\% for collision avoidance compared with TVIBLF-ECBF. The comparison between NN-TVIBLF-ECBF and NN-TVIBLF-AECBF shows that the maximum displacement in task space for collision avoidance is reduced by 35\%. The detailed tracking error for the x-, y- and z-axes are plotted in  Fig. \ref{fig:trackerrorstatic}. The given results show the application of a friction compensator reduces the tracking error of path tracking, while the proposed AECBF minimizes the abrupt changes during collision avoidance.

\begin{table}[htbp]
  \centering
  \caption{Maximum tracking errors of the path tracking task in scenario 1.}
   \small\addtolength{\tabcolsep}{-5pt}
  \resizebox{\linewidth}{!}{%
    \begin{tabular}{|c|c|c|c|}
    \hline
    Errors (m) & TVIBLF-ECBF & NN-TVIBLF-ECBF & NN-TVIBLF-AECBF \\
    \hline
    $|e_x|$  & 1.83E-02 & 3.66E-03 & $\bm{1.09E-03}$ \\
    \hline
    $|e_y|$  & 6.92E-03 & 2.12E-03 & $\bm{2.07E-03}$ \\
    \hline
    $|e_z|$  & 4.17E-03 & 2.60E-03 & $\bm{1.72E-03}$ \\
    \hline
    $|\Delta d|$  & 1.97E-02 & 4.99E-03 & $\bm{2.92E-03}$ \\
    \hline
    \end{tabular}}%
  \label{tab:eAstatic}%
\end{table}%

\begin{figure}
    \centering
     \subfloat[]{
    \includegraphics[width=0.45\textwidth]{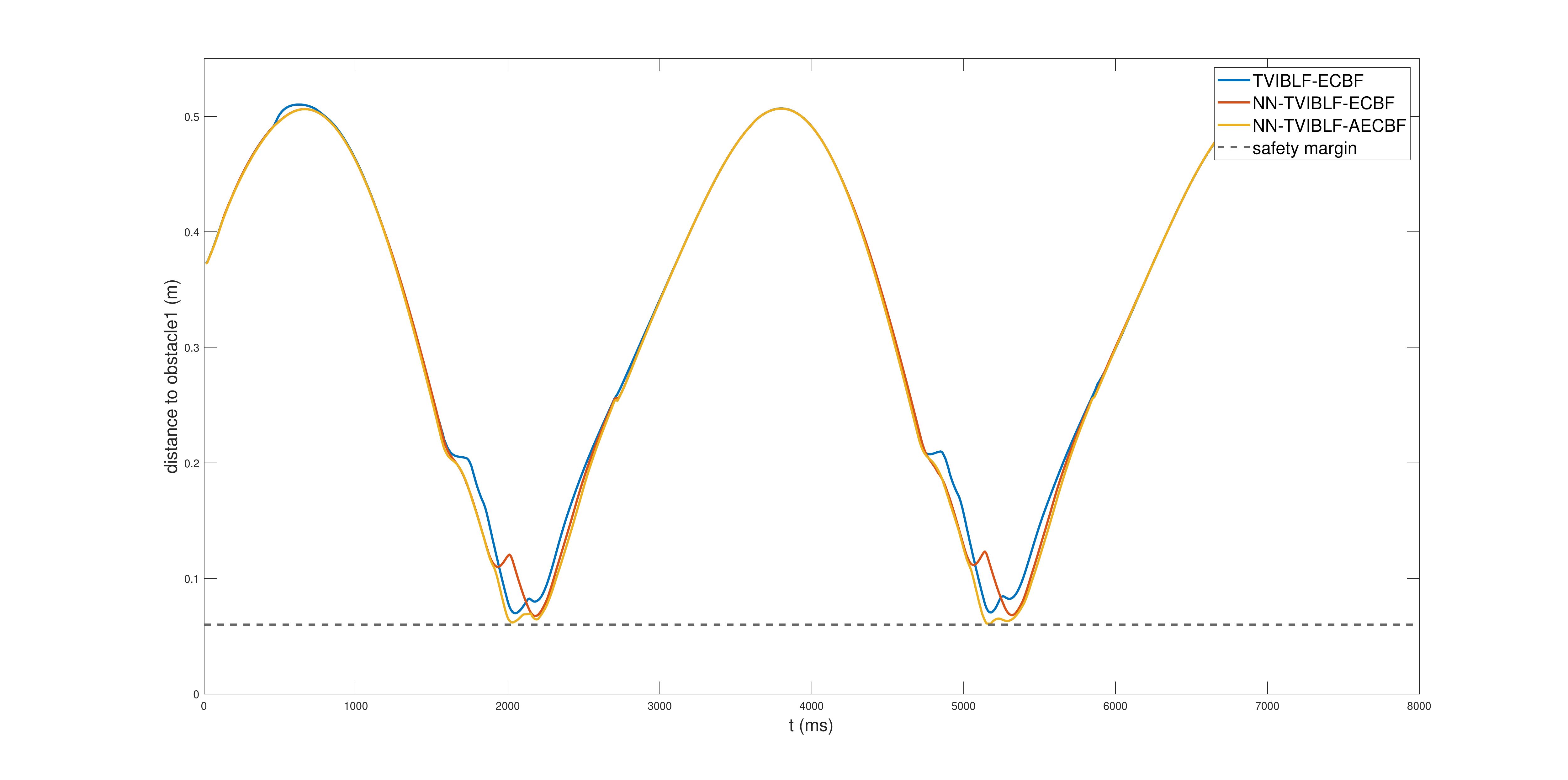}}
    
     \subfloat[]{
     \includegraphics[width=0.45\textwidth]{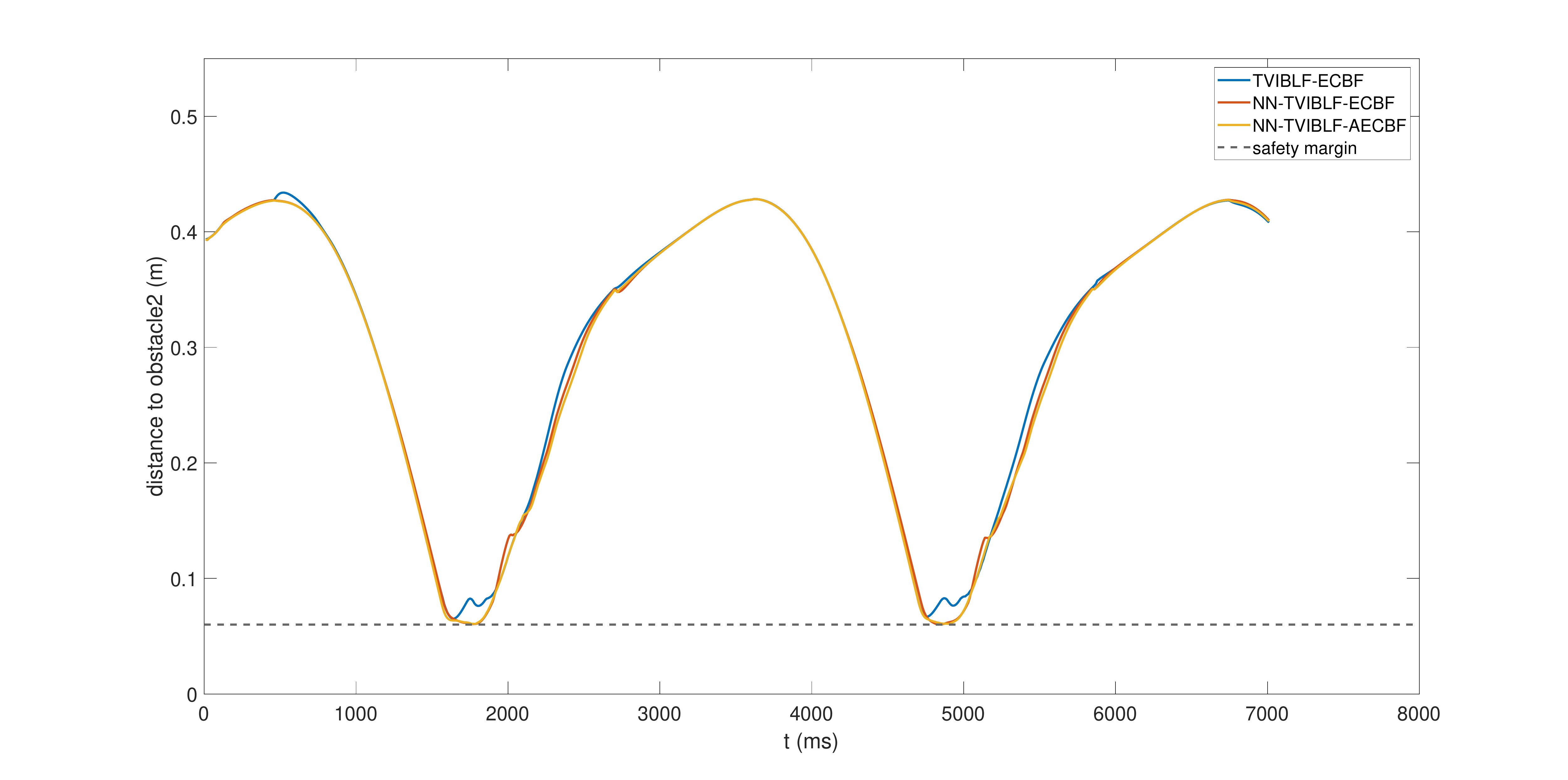}}
    \caption{Euclidean distance from the end-effector to both obstacles during HRC.}
    \label{fig:distancecase1}
\end{figure}

\begin{table}[htbp]
  \centering
  \caption{Maximum tracking errors of the collision avoidance task in scenario 1.}
   \small\addtolength{\tabcolsep}{-5pt}
  \resizebox{\linewidth}{!}{%
    \begin{tabular}{|c|c|c|c|}
    \hline
    Errors(m) & TVIBLF-ECBF & NN-TVIBLF-ECBF & NN-TVIBLF-AECBF \\
    \hline
    $|e_x|$  & 0.098 & 0.118 & $\bm{0.080}$ \\
    \hline
    $|e_y|$  & 0.035 & 0.035 & $\bm{0.024}$ \\
    \hline
    $|e_z|$  & 0.016 & 0.007 & $\bm{0.007}$ \\
    \hline
    $|\Delta d|$  & 0.11 & 0.13 & $\bm{0.084}$ \\
    \hline
    \end{tabular}}%
  \label{tab:eBstatic}%
\end{table}%
In order to check the state of collision with static obstacles, we plot the minimum distance between the robot's end-effector and obstacles in Fig. \ref{fig:distancecase1}. Fig. \ref{fig:distancecase1}(a) and (b) plot the minimum distance between the robot's end-effector with sphere 1 and sphere 2 separately. The dotted line represents the safety margin which is computed as the sum of the radii of the obstacle (0.05) and the safe distance (0.01). It shows that the NN-IBLF-AECBF model is able to obtain the minimum distance of 0.01 m to obstacles during collision avoidance, while the IBLF-ECBF and NN-IBLF-ECBF both have a safe distance of 0.012 m. 

\begin{figure}
    \centering
    \subfloat[]{
    \includegraphics[width=0.5\textwidth]{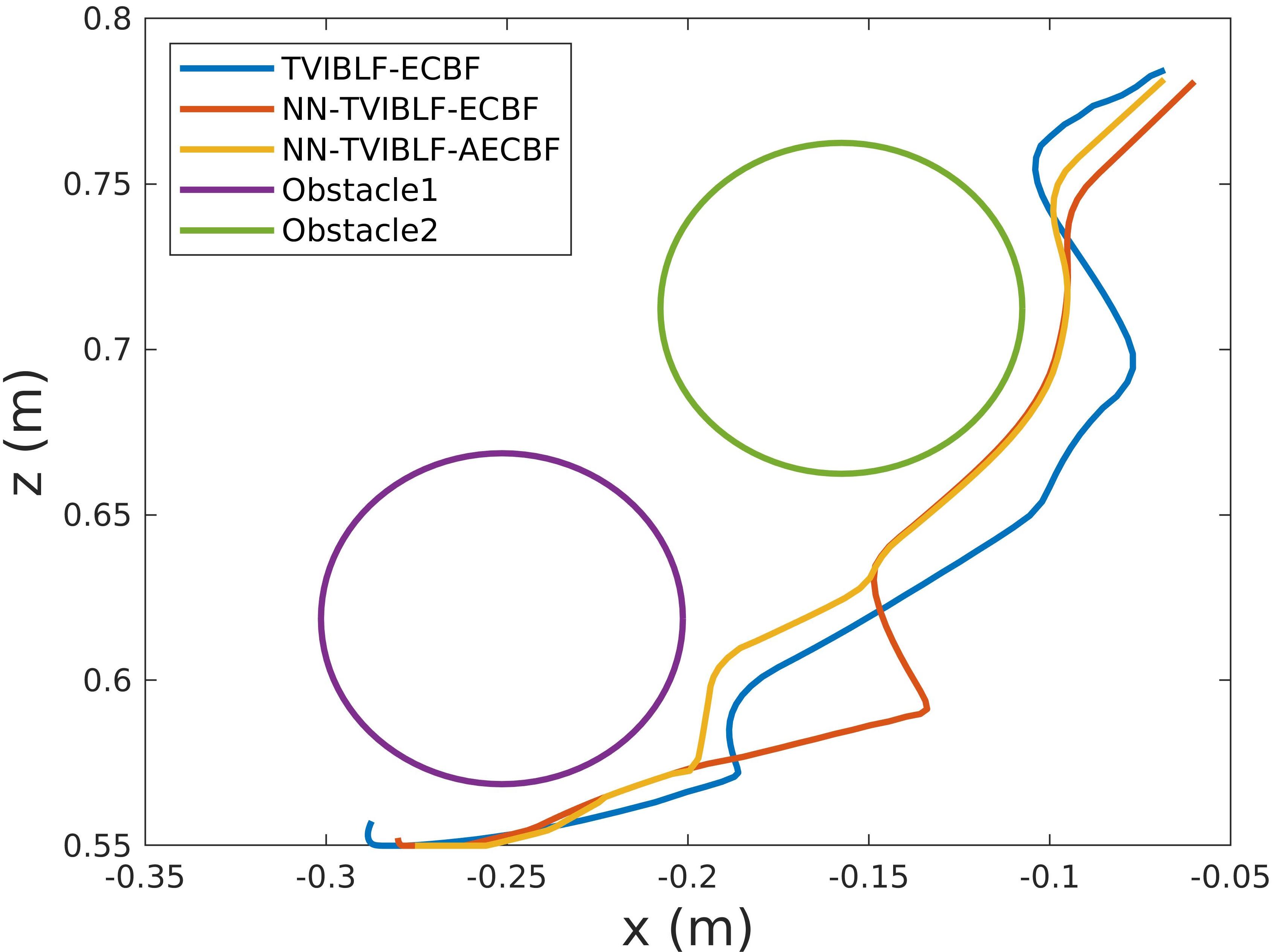}}
    
    \subfloat[]{
    \includegraphics[width=0.35\textwidth, height= 0.5\textwidth]{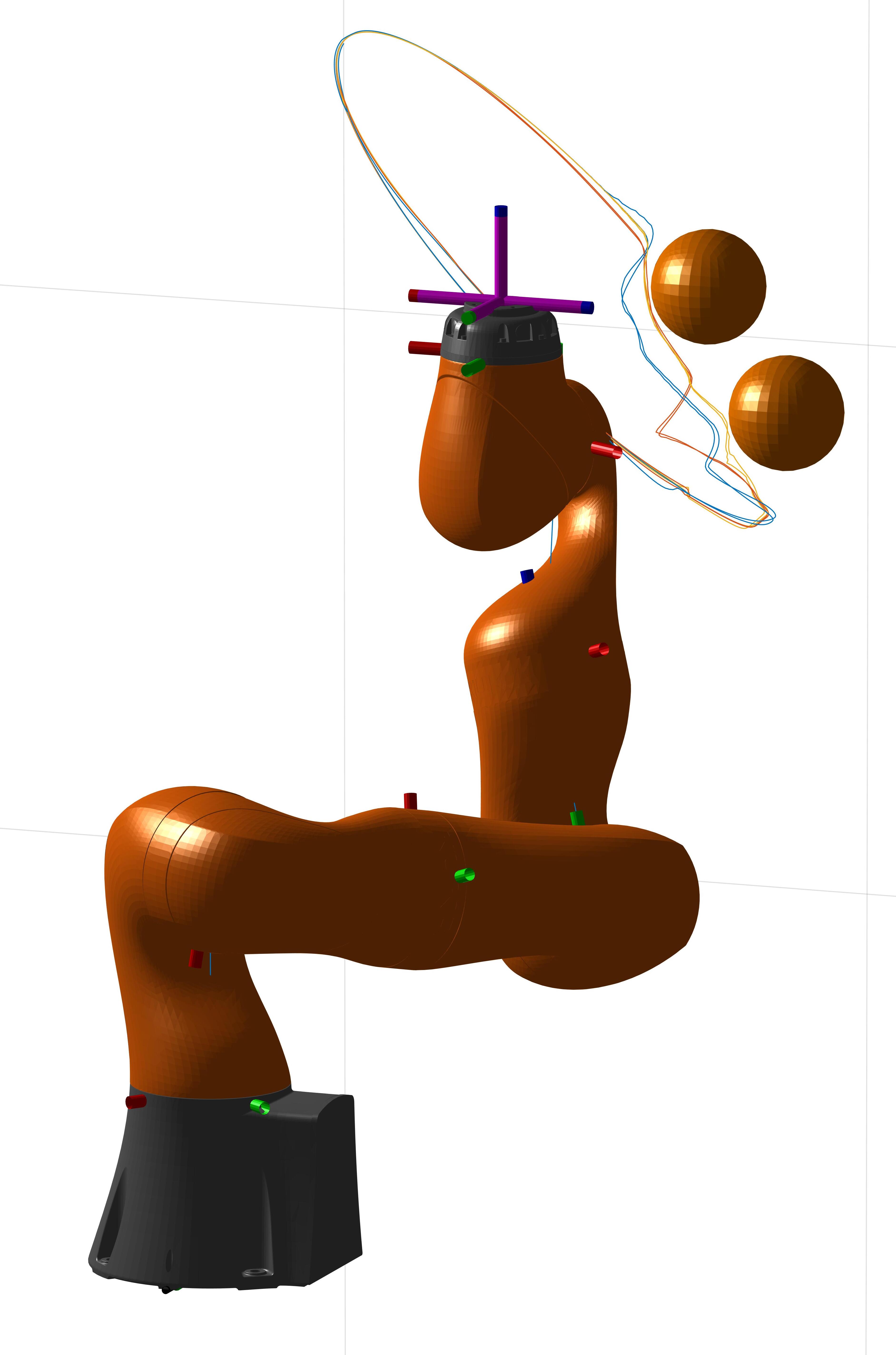}}
    \caption{Collision avoidance with static obstacles on Kuka LBR in (a) 2D and; (b) 3D environments.}
    \label{fig:2D3Dplot}
\end{figure}

The visualization of the trajectories generated by the three different models in 2D and 3D perspectives are shown in Fig. \ref{fig:2D3Dplot} (a) and (b), respectively. Comparing the trajectories for the controller with (red) and without (blue) RBFNN uncertainty estimation, it can be observed that uncertainty estimation enables smoother collision avoidance. However, the red trajectory takes an unnecessarily large deviation to avoid the first obstacle. In contrast, the NN-IBLF-AECBF trajectory (yellow) achieves the minimum deviation from the obstacle free path with fewer abrupt changes while satisfying the safety constraints.

\subsection{Case2: Collision Avoidance with Dynamic Obstacles}
\begin{figure} 
\begin{subfigure}{0.3\textwidth}
\includegraphics[width=0.5\linewidth]{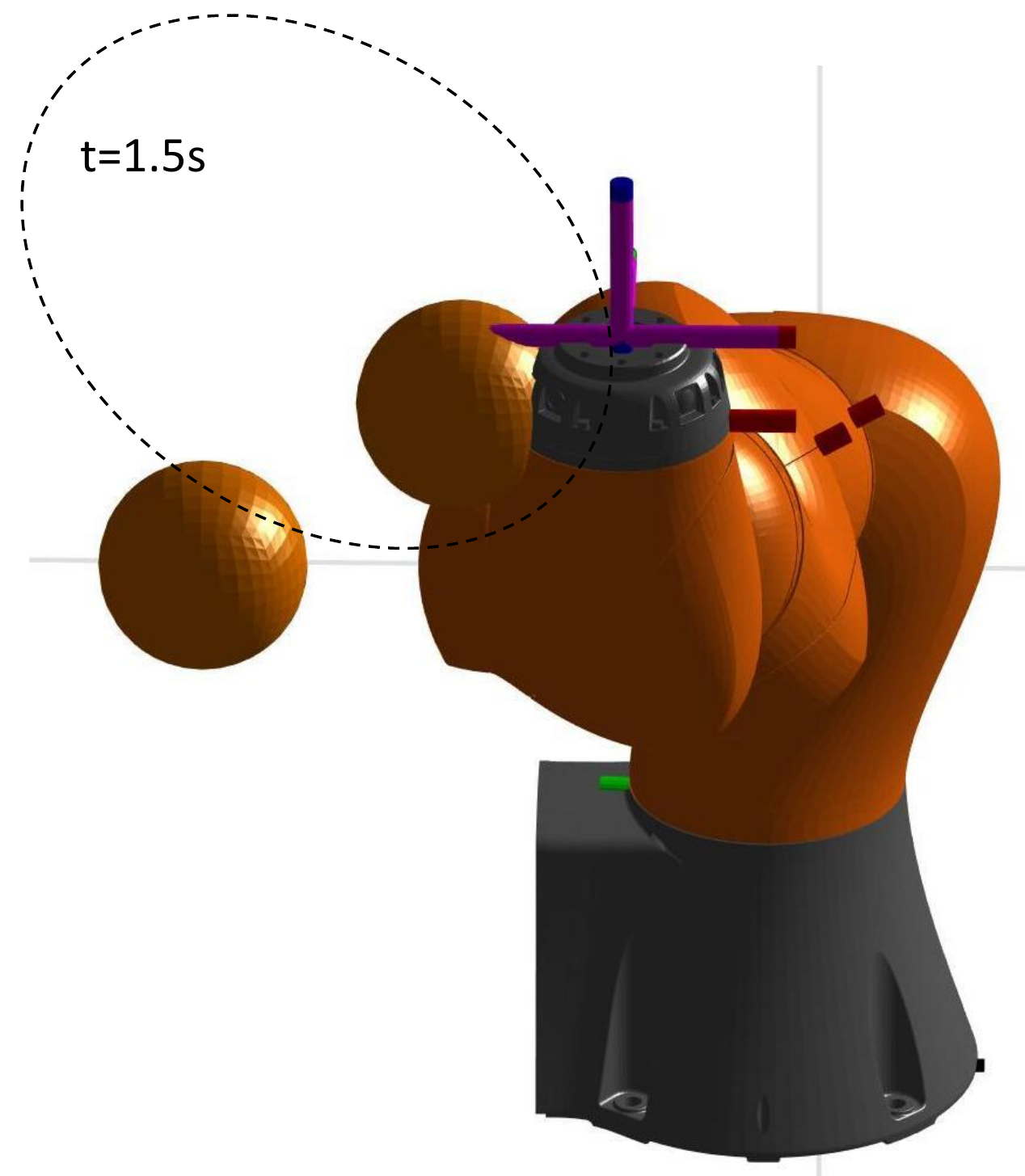}
\subcaption[]{}
\end{subfigure}\hspace*{\fill}
\begin{subfigure}{0.3\textwidth}
\includegraphics[width=0.5\linewidth]{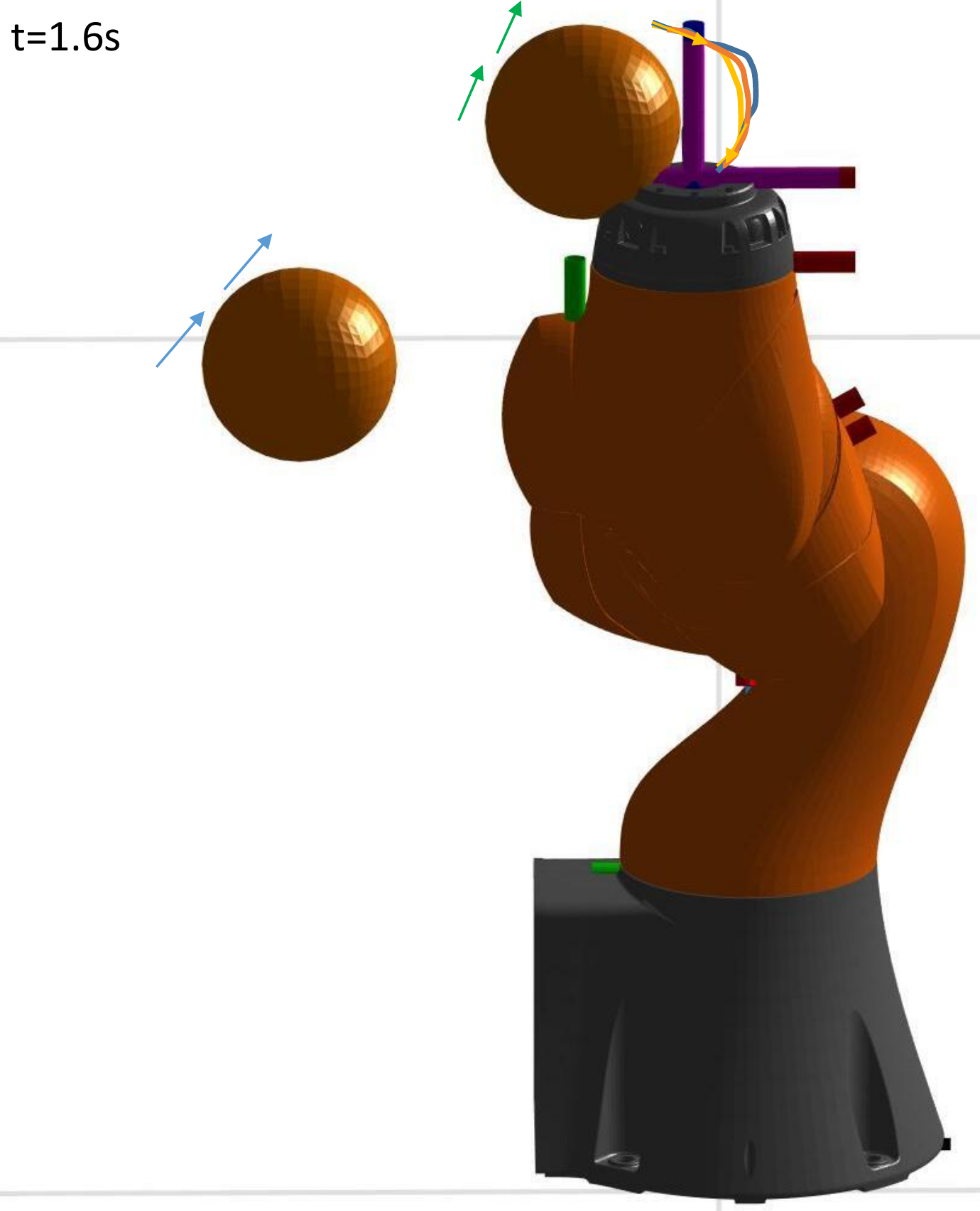}
\subcaption[]{}
\end{subfigure}
\begin{subfigure}{0.3\textwidth}
\includegraphics[width=0.5\linewidth]{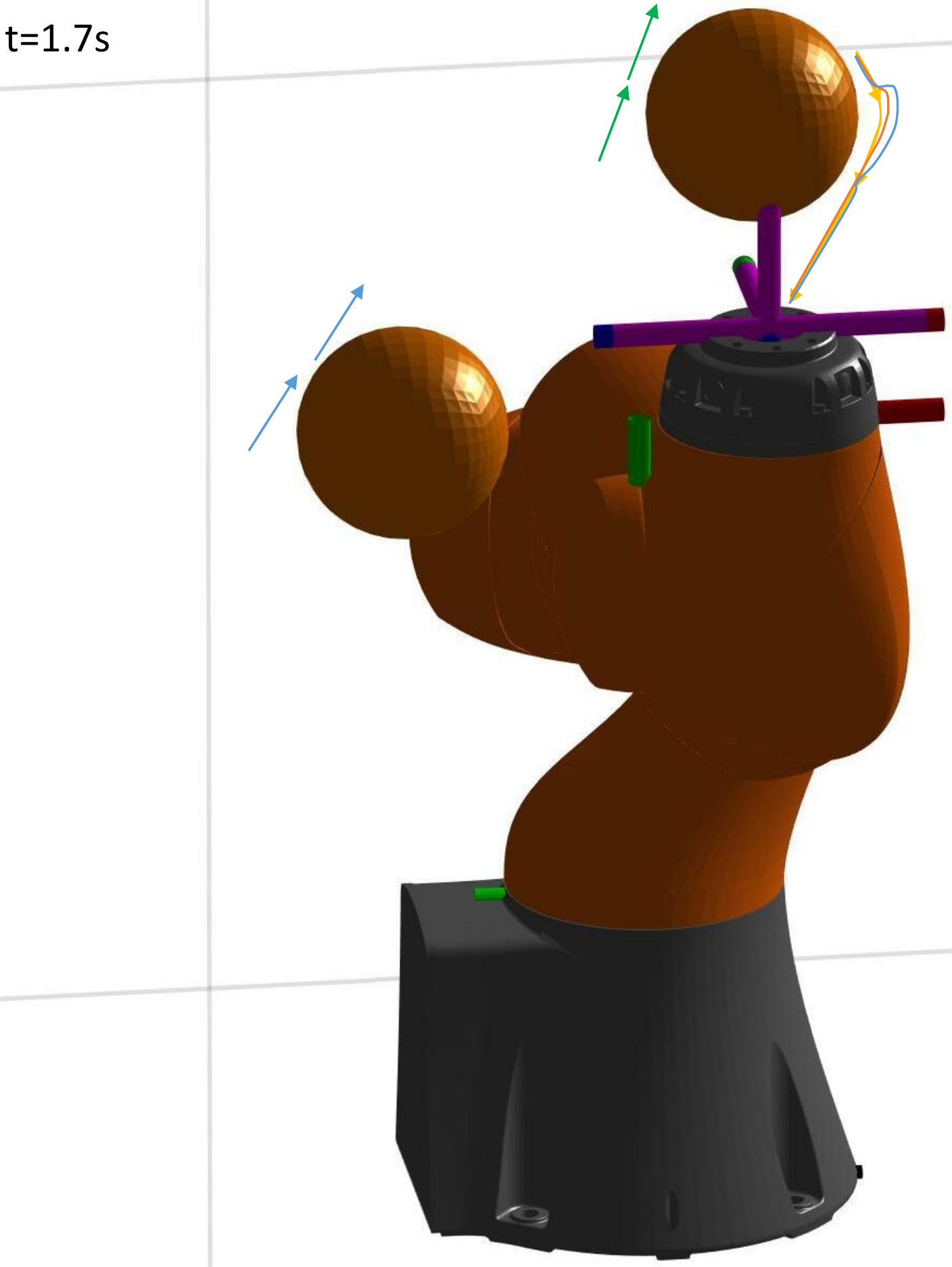}
\subcaption[]{}
\end{subfigure}\hspace*{\fill}
\begin{subfigure}{0.3\textwidth}
\includegraphics[width=0.5\linewidth]{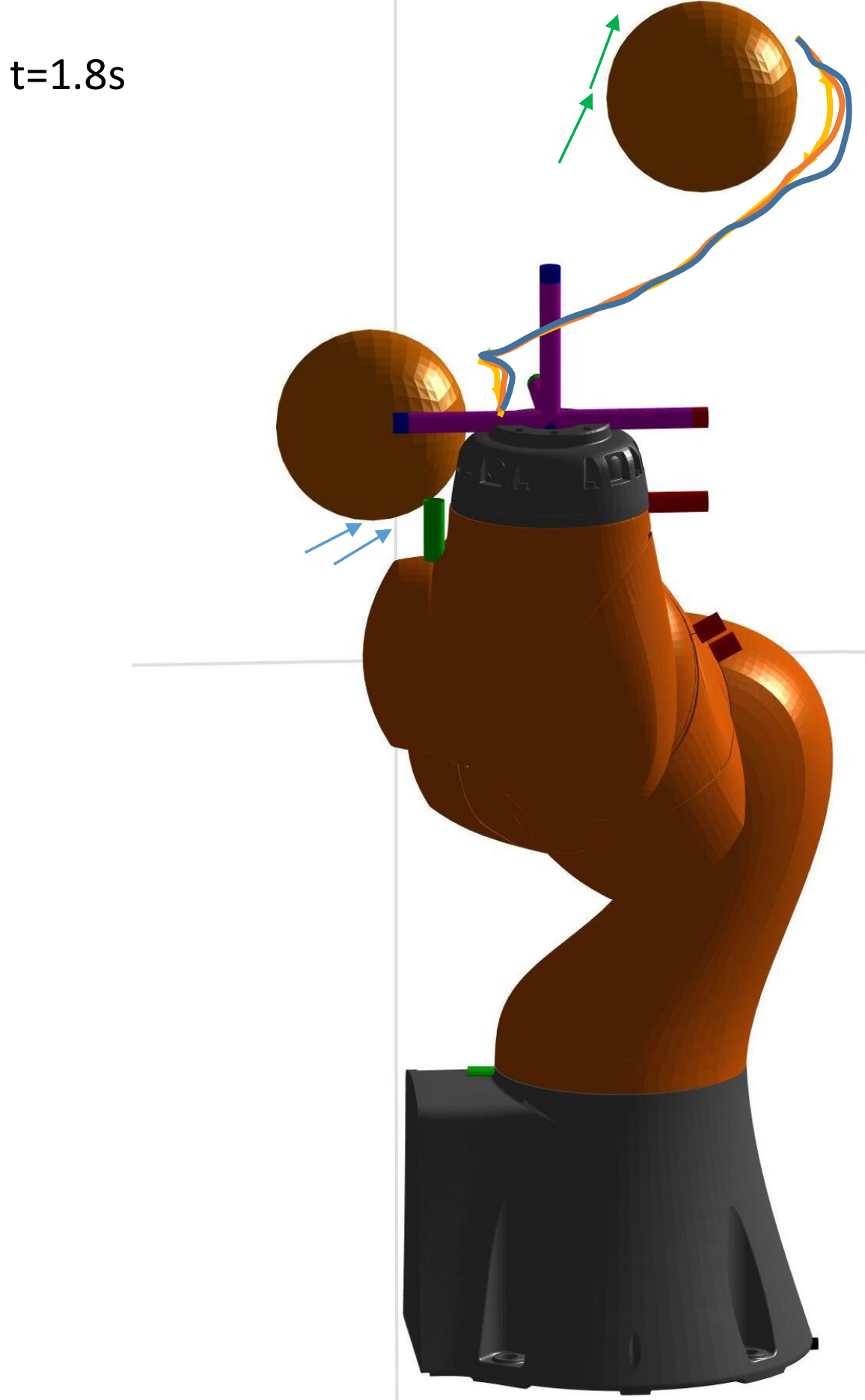}
\subcaption[]{}
\end{subfigure}
\begin{subfigure}{0.3\textwidth}
\includegraphics[width=0.5\linewidth]{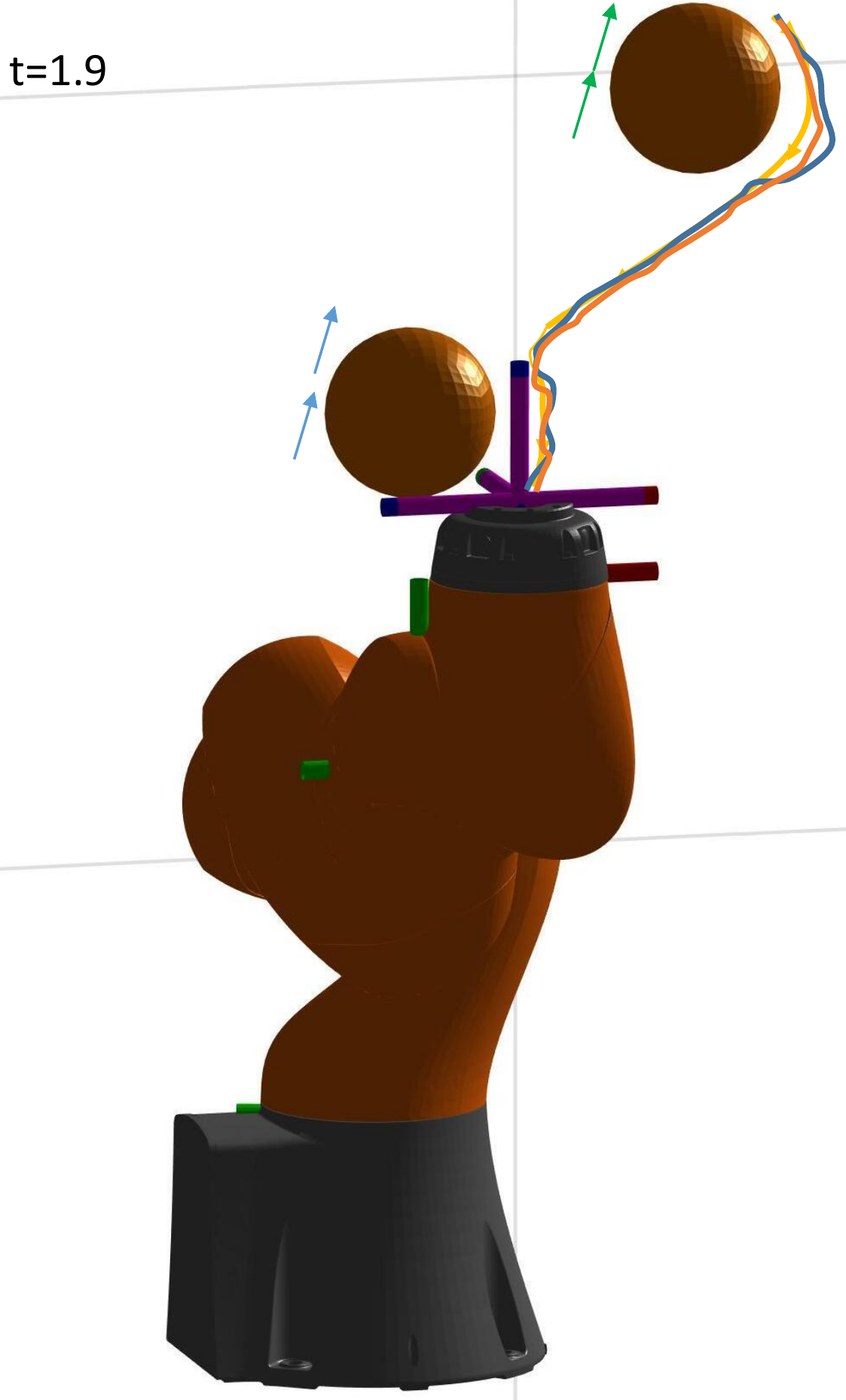}
\subcaption[]{}
\end{subfigure}\hspace*{\fill}
\begin{subfigure}{0.3\textwidth}
\includegraphics[width=0.5\linewidth]{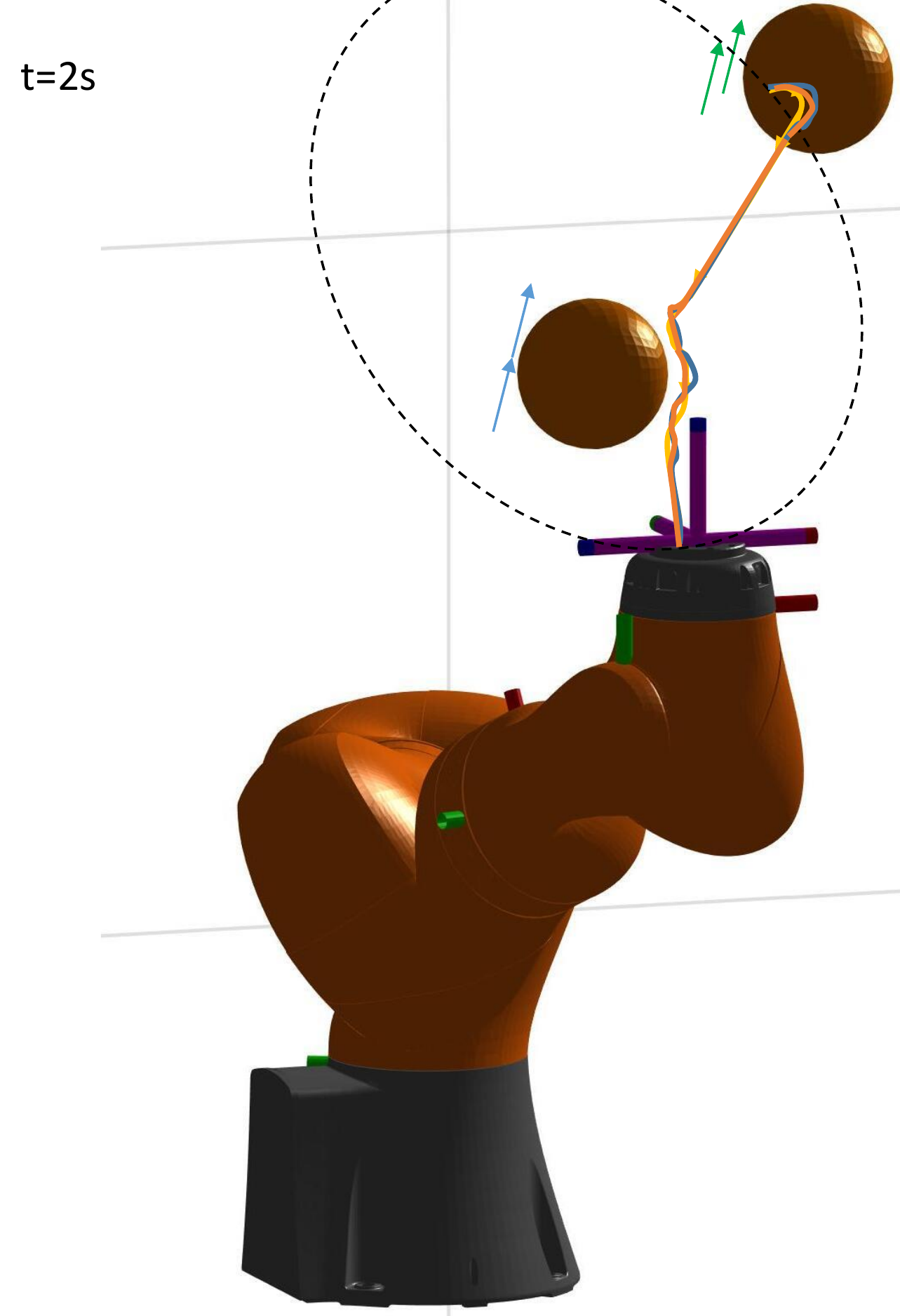}
\subcaption[]{}
\end{subfigure}
\caption{
Collision avoidance with NN-IBLF-NN-ECBF under dynamic environments. (a) The robot is tracking the desired trajectory (the dotted circle in Fig. \ref{fig:Case2trajectories} (a) and (f)). (b) The robot detects the first moving obstacle and performs collision avoidance. (c) The robot avoids collision with the first obstacle and plans to move toward the desired trajectory. (d) The robot detects the second moving obstacle and switches back to the collision avoidance task. (e) The robot finishes the collision avoidance with the second obstacle. (f) The robot moves back to the desired trajectory.}
\label{fig:Case2trajectories}
\end{figure}

\begin{figure}
     \centering
    \subfloat[]{\includegraphics[width=0.9\linewidth]{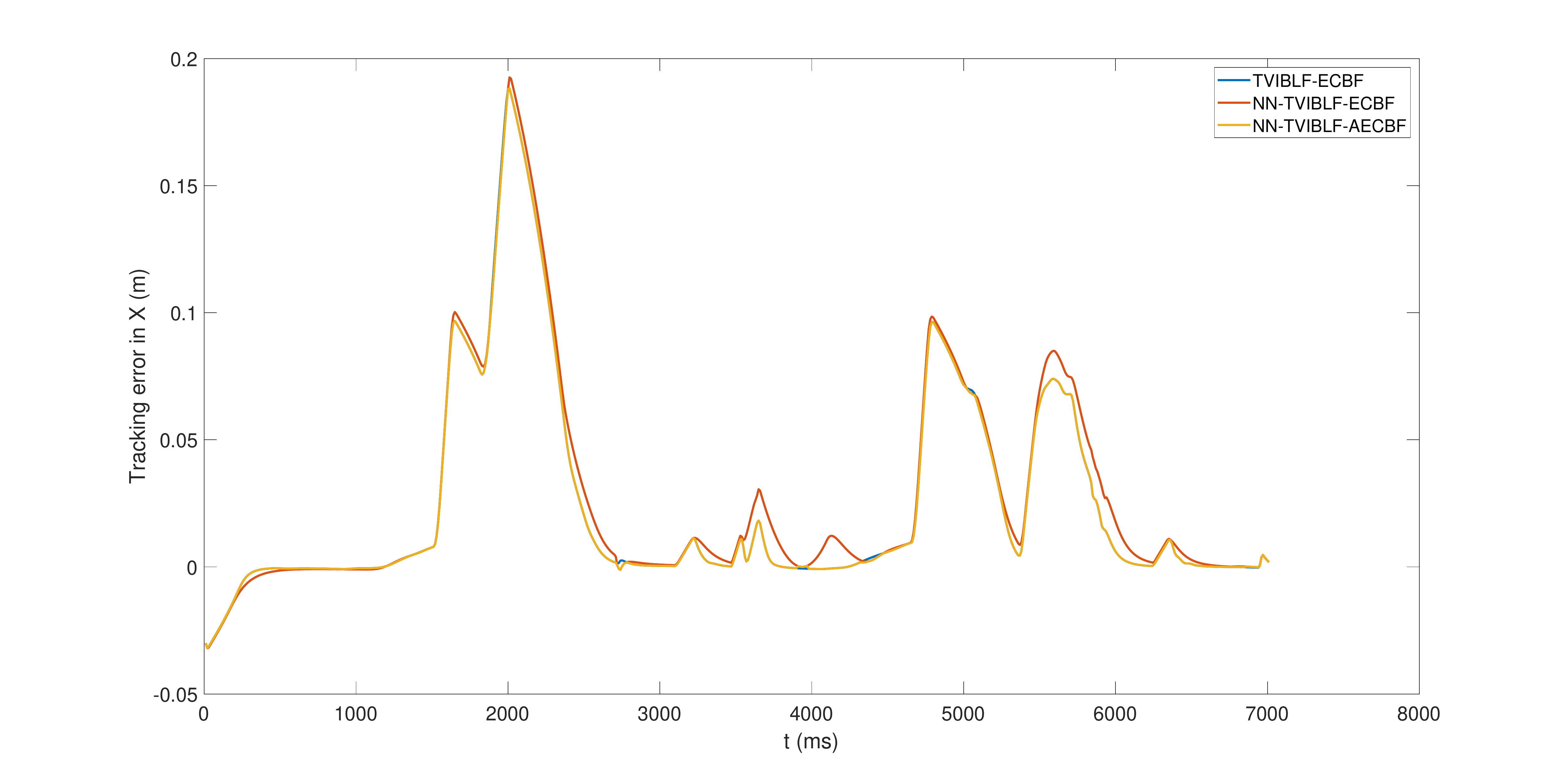}}\hfil
    \label{ob1}
    \subfloat[]{\includegraphics[width=0.9\linewidth]{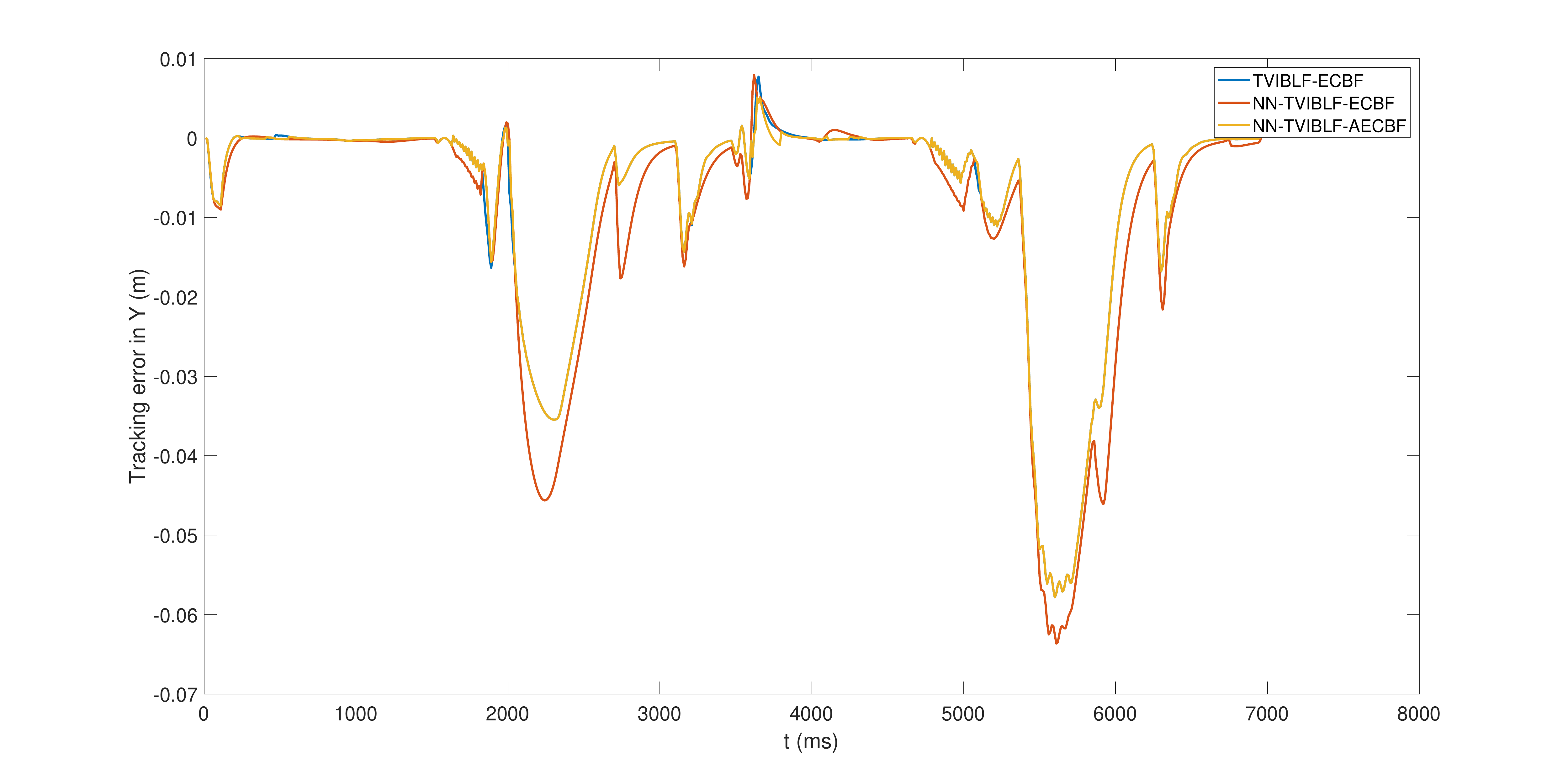}}\hfil
    \label{ob2}
    \subfloat[]{\includegraphics[width=0.9\linewidth]{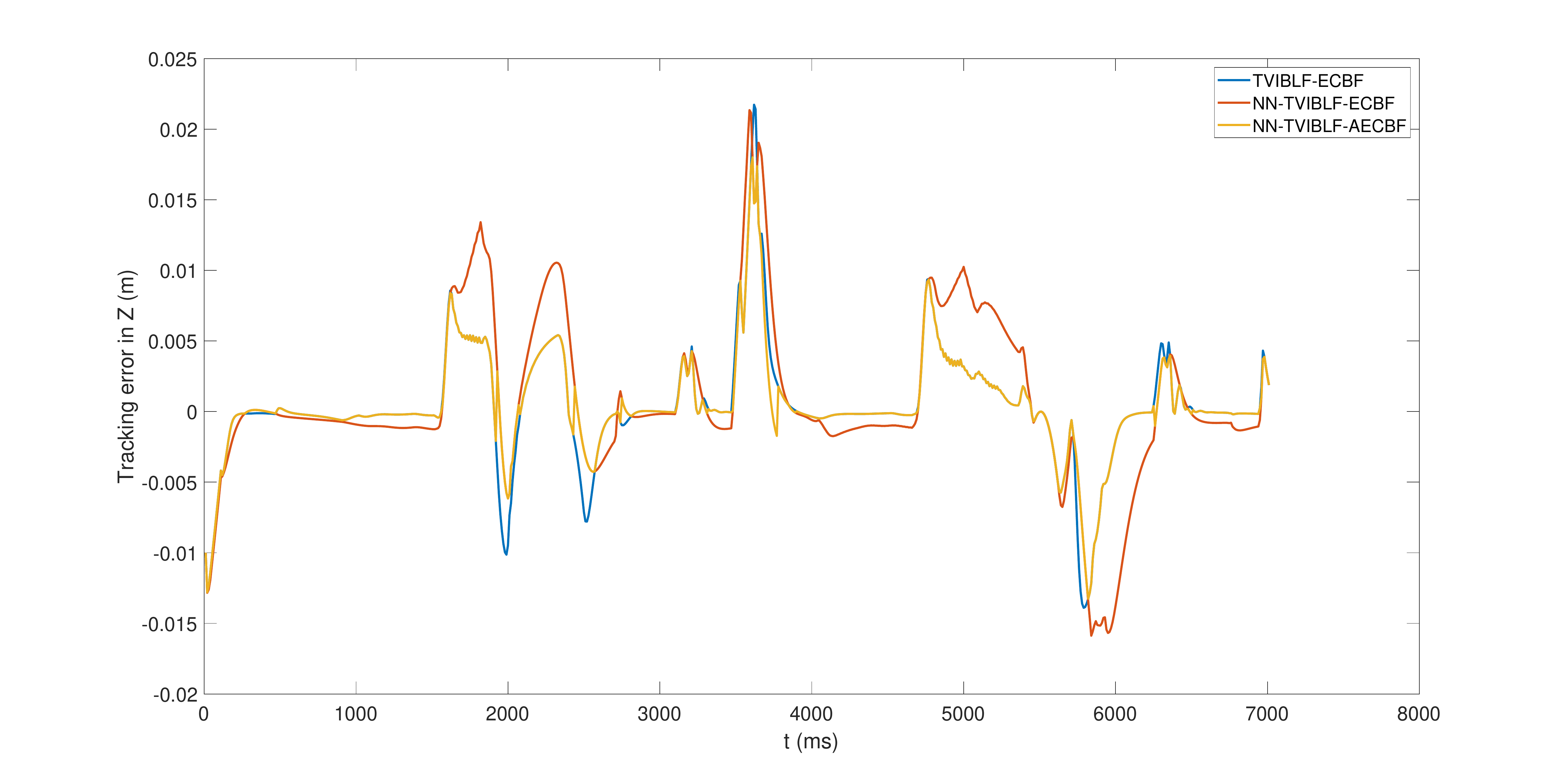}} 
     \caption{Comparison of the  trajectory tracking errors along the X, Y, and Z axes for scenario 2.}
     \label{fig:case2dyerror}
 \end{figure}

In this section, the robot manipulator follows the desired trajectory and simultaneously interacts with dynamic obstacles. The trajectory of the first obstacle is described by
\begin{align}
    \begin{split}
    &x_{d_x}(t) = (0.2sin(-1.5t)-0.1)(m),\\
    &x_{d_y}(t) = (0.2cos(-1.5t)-0.53)(m),\\
    &x_{d_z}(t) = (0.2sin(-1.5t)+0.77)(m).\\
    \end{split}
\end{align}
The trajectory of the second obstacle is described by
\begin{align}
    \begin{split}
    &x_{d_x}(t) = (0.2sin(-2t)-0.1)(m),\\
    &x_{d_y}(t) = (0.2cos(-2t)-0.53)(m),\\
    &x_{d_z}(t) = (0.2sin(-2t)+0.77)(m).\\
    \end{split}
\end{align}
where the defined obstacles move at different speeds. For better visualization, in Fig. \ref{fig:Case2trajectories}, we divide the motion from $t = 1.5 s$ to $t = 2 s$ into 6 segments and plot the movement of the robot and the obstacle during each segment. This shows how the robot responds to moving obstacles using different models. With the ability to continuously detect the changes in the spatial position of obstacles, the robot path planner recomputes the path immediately when obstacles violate the safe area and regulates the distance to the obstacles to the predefined safe distance. Fig. \ref{fig:Case2trajectories}(a) plots the state at the time instant when the ECBF is triggered, while Fig. \ref{fig:Case2trajectories}(f) plots the state when the robot finishes the collision avoidance task and moves back to the desired trajectory.  
The tracking and collision avoidance error is plotted in Fig. \ref{fig:case2dyerror}. Considering the information in Table. \ref{tab:case2trackerror} and Table. \ref{tab:case2cverror}, 
the proposed NN-IBLF-AECBF approach achieves the minimum tracking error and abrupt changes in collision avoidance, with an 11\% reduction in path tracking and a 4\% reduction in collision avoidance compared to TVIBLF-ECBF.
 
\begin{figure}
    \centering
     \subfloat[]{
    \includegraphics[width=0.45\textwidth]{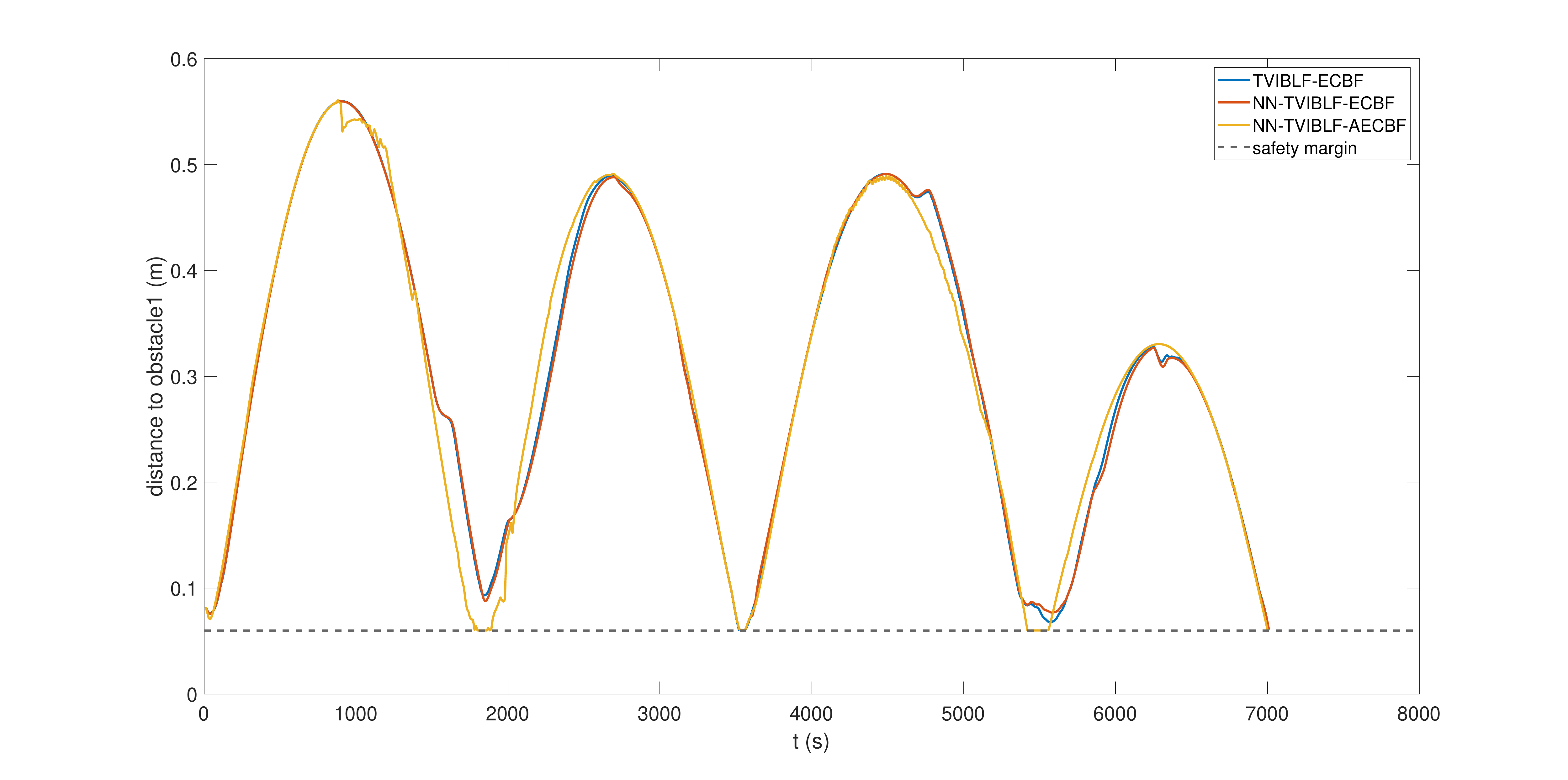}}
    
     \subfloat[]{
     \includegraphics[width=0.45\textwidth]{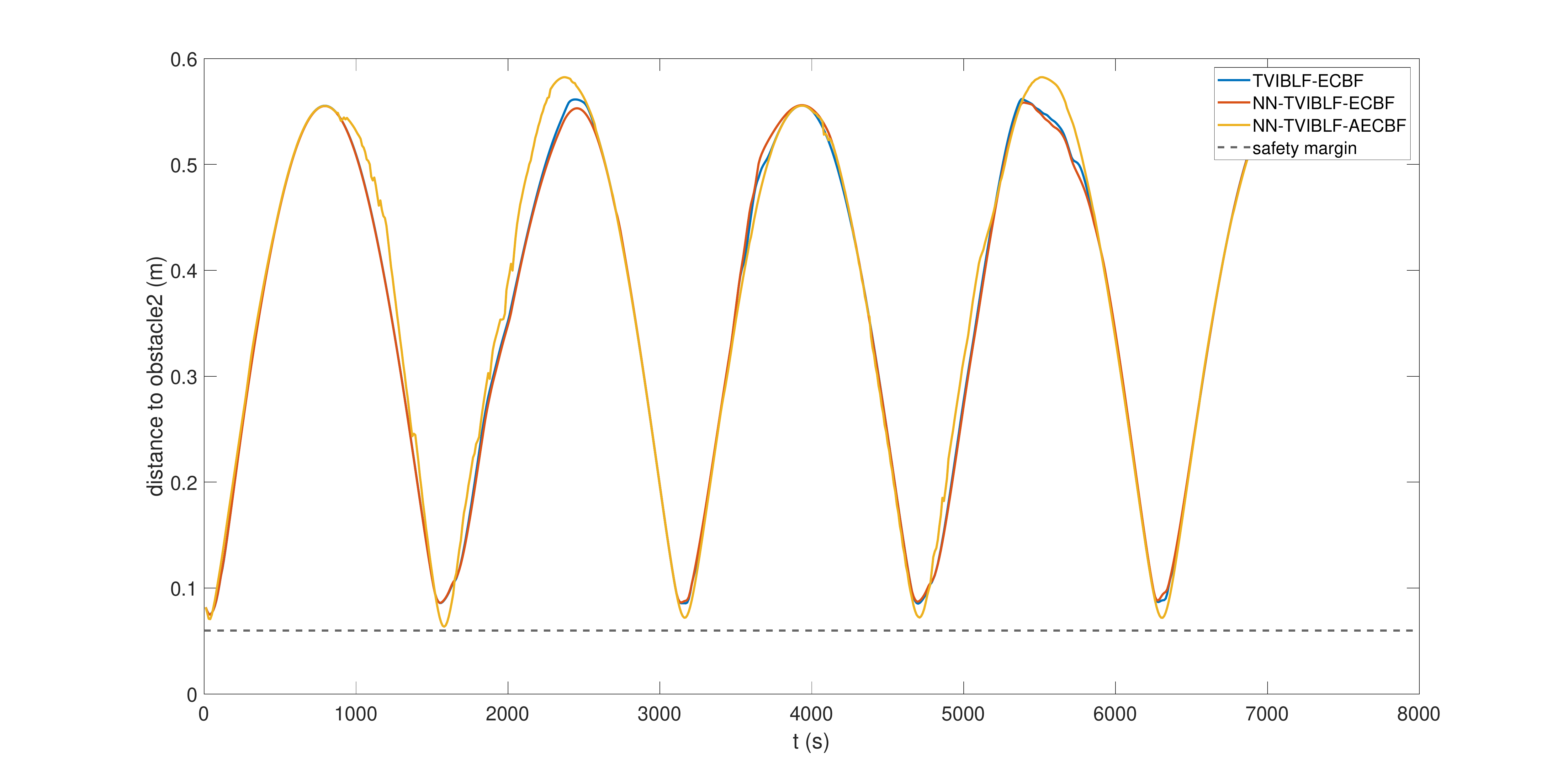}}
    \caption{Euclidean distance from the end-effector to both obstacles during HRC.}
    \label{fig:case2distance}
\end{figure}

 Fig.\ref{fig:case2distance} plots the control force for the end-effector of the KUKA LBR iiwa robot in three axes. For path tracking, the amplitude of the maximum control force is reduced from [24.3, 16.4, 14.7] N to [20.7, 15.1, 8.9] N. In the collision avoidance task, the amplitude of the maximum control force is reduced from [40, 17.3, 7.1] N to [30.2, 10.2, 4.9] N. It shows that the compensated system uncertainty helps to improve the performance of both path tracking and collision avoidance. Applying the AECBF in the collision avoidance task, further reduces the control force compared with NN-TVIBLF-ECBF, with a reduction of [2.7, 0.3, 0.4] N observed.
\begin{figure}
     \centering
    \subfloat[]{\includegraphics[width=0.9\linewidth]{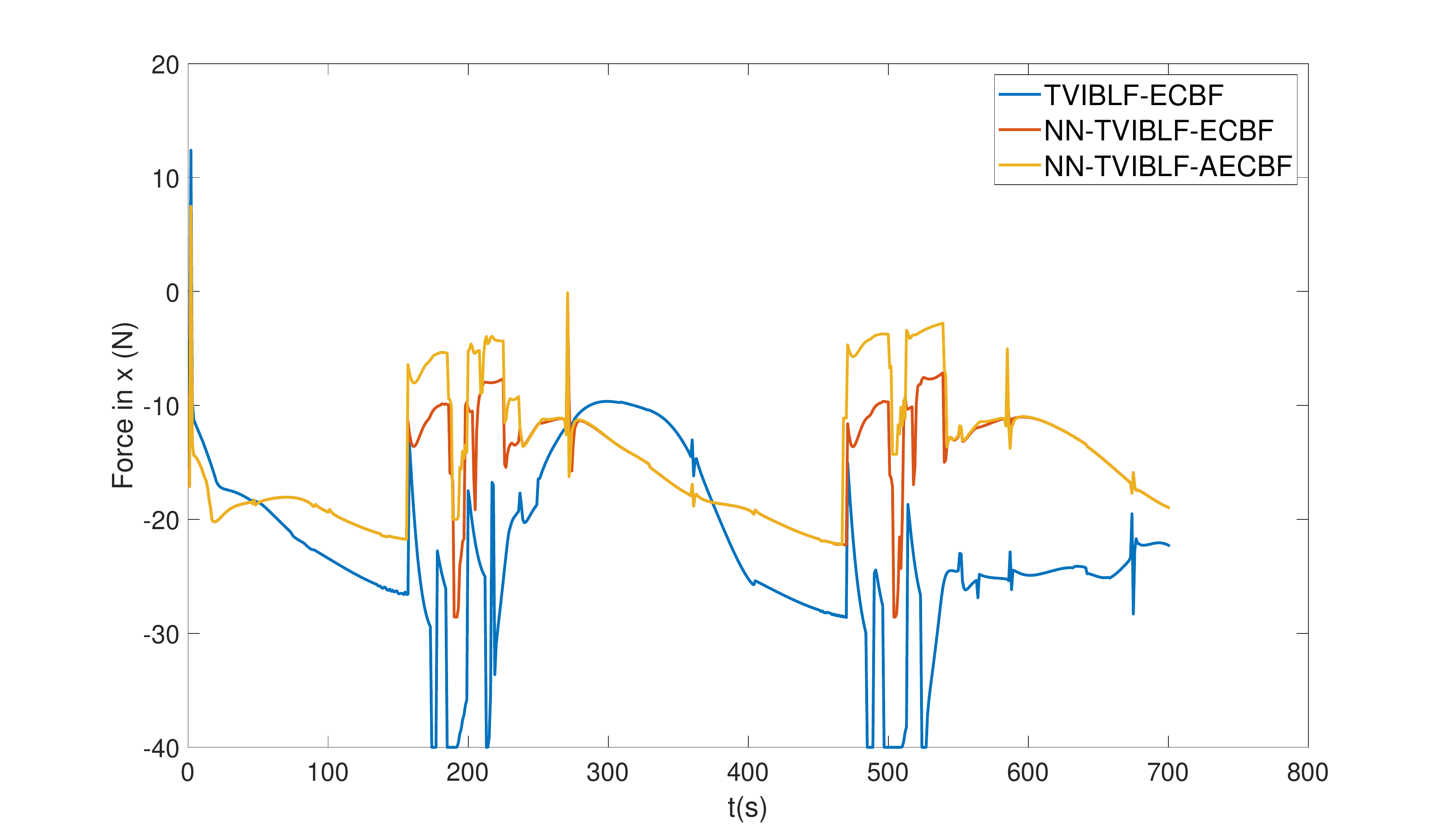}}\hfil
    \label{ob1}
    \subfloat[]{\includegraphics[width=0.9\linewidth]{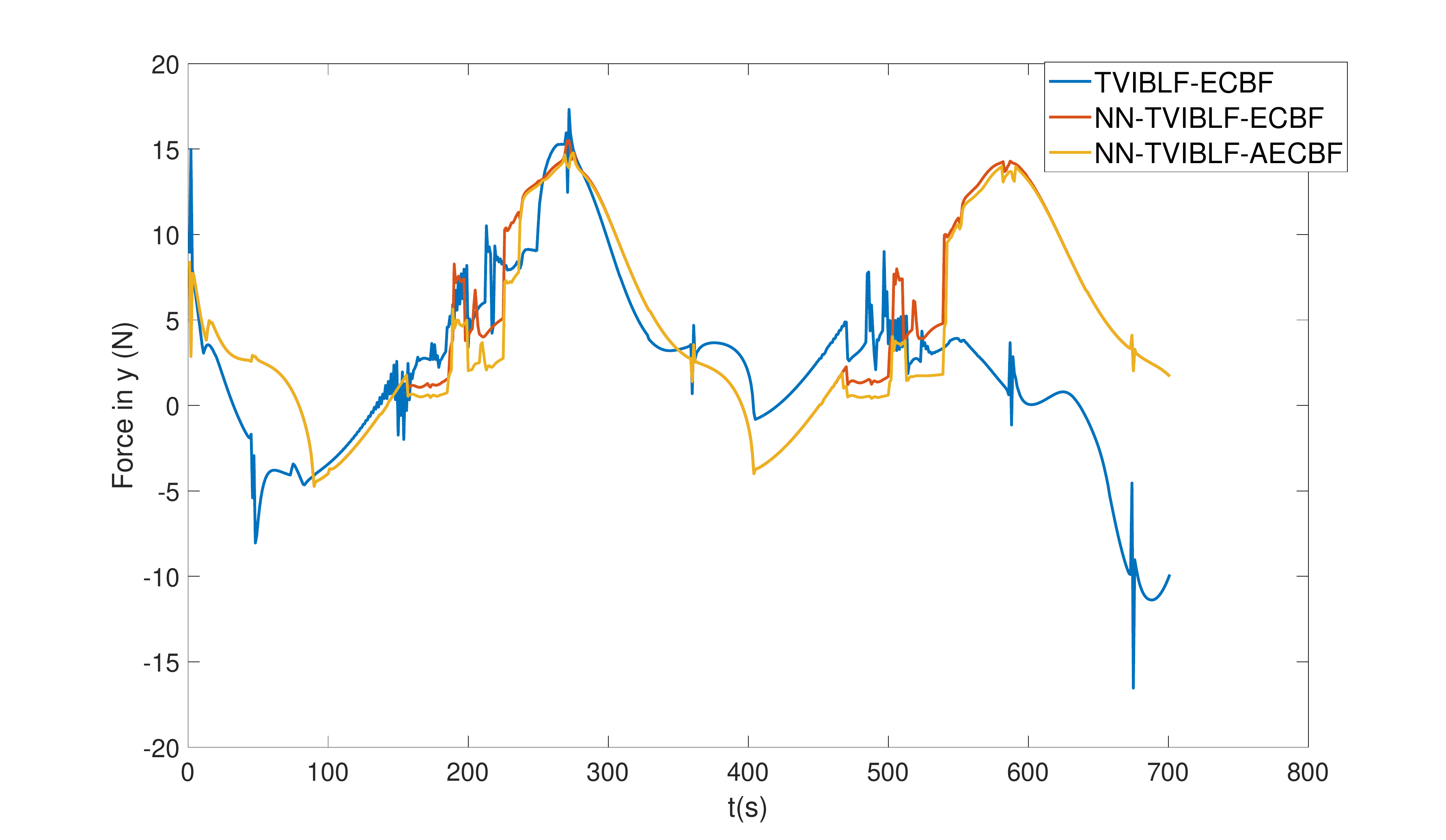}}\hfil
    \label{ob2}
    \subfloat[]{\includegraphics[width=0.9\linewidth]{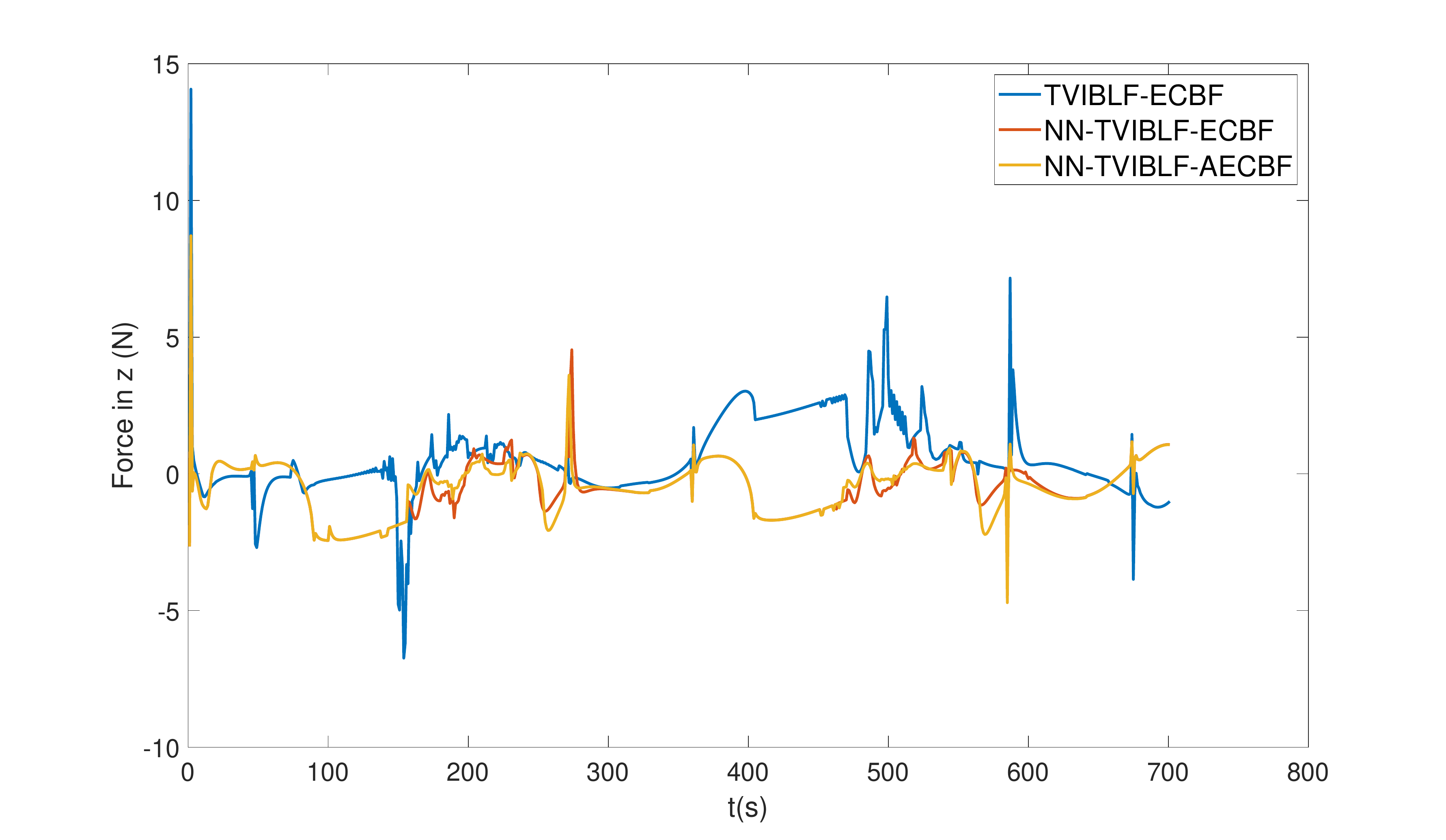}} 
     \caption{Comparison of the  control force along the X, Y, and Z axes for scenario 2.}
     \label{fig:case2dyforce}
 \end{figure}
 
\begin{table}[htbp]
  \centering
  \caption{Total movement and rotation to execute the designed path tracking and dynamic obstacle avoidance task on three models.}
   \small\addtolength{\tabcolsep}{-5pt}
  \resizebox{\linewidth}{!}{%
    \begin{tabular}{|c|c|c|}
    \hline
          & distance(m) & rotation(degree) \\
    \hline
    IBLF-ECBF & 4.69  & 971.2 \\
    \hline
    NN-IBLF-ECBF & 4.61  & 956.3 \\
    \hline
    NN-IBLF-NN-ECBF & $\bm{4.47}$  & $\bm{947.9}$ \\
    \hline
    \end{tabular}}%
  \label{tab:totaldynamic}%
\end{table}%

\begin{table}[htbp]
  \centering
  \caption{Maximum errors of path tracking in scenario 2.}
   \small\addtolength{\tabcolsep}{-5pt}
  \resizebox{\linewidth}{!}{%
    \begin{tabular}{|c|c|c|c|}
    \hline
    Errors(m) & TVIBLF-ECBF & NN-TVIBLF-ECBF & NN-TVIBLF-AECBF \\
    \hline
    $|e_x|$  & 2.57E-02 & 2.35E-02 & $\bm{2.22E-02}$ \\
    \hline
    $|e_y|$  & 3.53E-02 & 3.32E-02 & $\bm{3.17E-02}$ \\
    \hline
    $|e_z|$  & 2.51E-04 & 3.30E-04 & $\bm{3.22E-04}$ \\
    \hline
    $|\Delta d|$  & 0.044 & 0.052 & $\bm{0.039}$ \\
    \hline
    \end{tabular}}%
  \label{tab:case2trackerror}%
\end{table}%

\begin{table}[htbp]
  \centering
  \caption{Maximum errors of collision avoidance in scenario 2.}
   \small\addtolength{\tabcolsep}{-5pt}
  \resizebox{\linewidth}{!}{%
    \begin{tabular}{|c|c|c|c|}
    \hline
    Errors(m) & TVIBLF-ECBF & NN-TVIBLF-ECBF & NN-TVIBLF-AECBF \\
    \hline
    $|e_x|$  & 0.187 & 0.187 & $\bm{0.182}$ \\
    \hline
    $|e_y|$  & 0.064 & 0.064 & $\bm{0.057}$ \\
    \hline
    $|e_z|$  & 0.022 & 0.021 & $\bm{0.017}$ \\
    \hline
    $|\Delta d|$  & 0.199 & 0.199 & $\bm{0.191}$ \\
    \hline
    \end{tabular}}%
  \label{tab:case2cverror}%
\end{table}%
Fig. \ref{fig:case2distance} plots the minimum distance between the robot and the moving obstacles, where no collision happens when the robot interacts with the dynamic obstacles. It shows that NN-TVIBLF-AECBF helps to implement collision avoidance with a closer distance to the safe boundary, while the other two controllers keep a further distance from the moving obstacles. The results in Table \ref{tab:totaldynamic} and \ref{tab:case2cverror} show that the proposed AECBF improves the performance of collision avoidance by obtaining the minimum total movement and joint rotations.

\section{Conclusion}
In this work, the uncertainty in the robot dynamics learned by an RBFNN is propagated to SCC-based path tracking and collision avoidance approaches to provide accurate positioning of the robot. Additionally, a machine learning-based method is adopted to improve the performance of the AECBF module for shortest path selection. The results demonstrate the benefits of adding a model uncertainty compensator and employing an AECBF-based safety filter in HRC tasks, in terms of guaranteeing the safety of human operators and improving the efficiency of HRC tasks even in cases of deliberate human constraint violation. 

The proposed controller does have some limitations which need to be considered in real-world implementation. There may be a greater number of configurations of the robot and obstacles where a feasible solution does not exist with regard to satisfying both the TVIBLF and AECBF constraints, compared to other approaches -- the price of being more conservative to guarantee safety. Also, we employ a simple spherical representation for the obstacles to be avoided in the HRC task. While this is beneficial in terms of reducing computational complexity it is very conservative in terms of defining the acceptable operating space for robot motion. To mitigate these limitations, in future work we will look at combining our approach with a method that provides more precise detection and bounding of obstacles in the robot operating space (e.g. using LIDAR and semantic segmentation).

\printbibliography
\end{document}